\pdfoutput=1
\documentclass[review]{elsarticle}
\usepackage{subfigure}
\usepackage{longtable}
\usepackage{algorithm, algorithmic}
\usepackage{bm}
\usepackage{booktabs}
\usepackage{multirow}
\usepackage{geometry}
\usepackage{amsfonts,amssymb}
\usepackage{amsmath}
\usepackage{array}

\usepackage{lineno,hyperref}
\modulolinenumbers[5]

\journal{arXiv}

\begin{document}

\begin{frontmatter}

\title{Multi-Scale Iterative Refinement Network for RGB-D Salient Object Detection}


\author[mymainaddress]{Ze-yu Liu}

\author[mymainaddress]{Jian-wei Liu\corref{mycorrespondingauthor}}
\cortext[mycorrespondingauthor]{Corresponding author}
\ead{liujw@cup.edu.cn}

\author[mymainaddress]{Xin Zuo}

\author[mymainaddress]{Ming-fei Hu}

\address[mymainaddress]{Department of Automation
College of Information Science and Engineering
China University of Petroleum, Beijing Campus (CUP)
Beijing, China
}

\begin{abstract}
The extensive research leveraging RGB-D information has been exploited in salient object detection. However, salient visual cues appear in various scales and resolutions of RGB images due to semantic gaps at different feature levels. Meanwhile, similar salient patterns are available in cross-modal depth images as well as multi-scale versions. Cross-modal fusion and multi-scale refinement are still an open problem in RGB-D salient object detection task. In this paper, we begin by introducing top-down and bottom-up iterative refinement architecture to leverage multi-scale features, and then devise attention based fusion module (ABF) to address on cross-modal correlation. We conduct extensive experiments on seven public datasets. The experimental results show the effectiveness of our devised method. 
\end{abstract}


\end{frontmatter}

\section{Introduction}

Salient object detection is an important task in the field of computer vision due to its numerous applications in image matting 
\cite{1xue2013automatic}, semantic segmentation 
\cite{2han2018reinforcement}, product retrieval 
\cite{3wang2019retrieval}, and edge detection 
\cite{4hou2018three}. With current advances in deep convolutional neural networks (CNNs), salient object detection has been promoted greatly. Different from RGB salient object detection, RGB-D salient object detection aims to distinguish the most significant objects from less attractive background regions in RGB image with the aid of depth sensors. In addition to the RGB signal, depth sensor provides complementary cues for salient object detection. To effectively utilize RGB and depth information, a number of researches 
\cite{5piao2019depth,6zhao2019contrast,7chen2018progressively} have been proposed to integrate multi-modal visual cues. On the other hand, inspired by feature pyramid networks 
\cite{8lin2017feature}, several works 
\cite{9chen2019multi,10pang2020multi,11wang2018salient} propose to aggregate multi-scale features from high-level and low-level representations.
Despite promising improvements on feature representation brought by current advances, there still exists problems in the following aspects:

\textbf{Semantic gaps between hierarchical deep features with multiple scales.} As depicted in Fig. 1, the low-level features on the early semantic feature maps fire local visual cues such as edges or skeleton. However, the high-level layers with large scales tend to respond to semantic visual cues such as instances or instance parts, and the response maps from these layers usually represent global features. Obviously, there exists a semantic gap between hierarchical features. Specifically, low-level features work well in the edge detection task. By contrast, high-level features are useful for classification or detection task. How to bridge the semantic visual gaps between multi-scale hierarchical features is an open question.
\begin{figure}[!htbp]
	\centering
	\includegraphics[width=\textwidth]{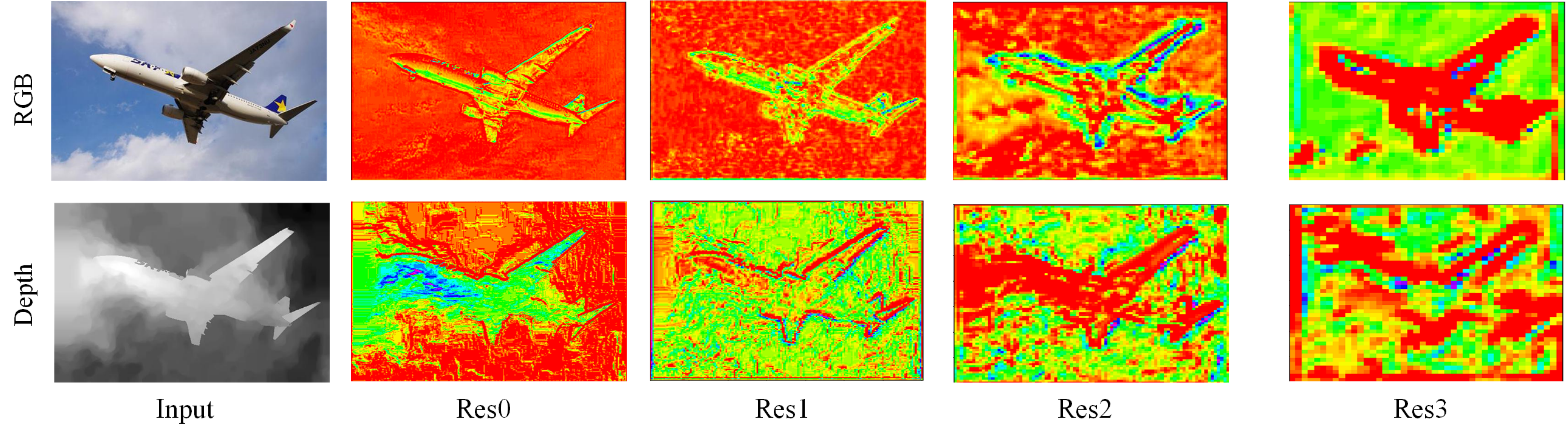}
	\caption{The response map of high-level and low-level semantic features from ResNet neural network. Both RGB and depth images are fed to neural network. Subplots from left to right are selected from convolutional layers res0 to res3. Best visualization in color.}
	\label{fig1}
\end{figure}

\textbf{Information loss of the high-level feature map.}Salient object detection, as researched in enormous research works 
\cite{12hou2017deeply,13li2016deep,14li2017instance}, requires the ability to extract the most conspicuous objects which are based on extracting high-level features. High-level features in the convolutional neural network contain affluent information, which are beneficial to locate conspicuous objects. However, in common detection protocol, visual signals propagate in a top-down manner, This can result in lossing fine-grained local features of objects.

\textbf{Complementary information mining in cross-modal features.} RGB image contains rich semantic information in salient object detection. On the other hand, corresponding depth maps contain abundant spatial and layout features, which provide geometrical guidance to improve the accuracy of salient object detection. Obviously, it is non-trivial to mine and integrate information from two different modals. To this end, various approaches 
\cite{15li2014saliency,6zhao2019contrast} have been proposed to solve the problem. Current methods propose to use early fusion 
\cite{16peng2014rgbd,17song2017depth}, late fusion 
\cite{18cheng2014depth} or middle fusion 
\cite{19feng2016local} to mine correlation in RGB and depth features. However, existing methods still lack considering the mutual relationship between cross-modal data.

To address on these issues, we propose a novel network MSIRN, which uses both multi-scale and cross-modal features to reinforce the performane for salient object detection. Our scheme contains iterative refinement module with both top-down and bottom-up fashion. The intuition behind our approach is based on the idea that iterative refinement could bridge the semantic gaps between multi-scale features. Top-down and bottom-up feature propagation mitigate the information loss of high-level semantic features. Besides, attention-based cross-modal correlation is leveraged to mine complementary information between paired RGB-D images.

In short, our contributions can be summarized as follows:

1)	We define an iterative refinement schema to generate coarse-to-fine saliency maps. In addition, we further employ both bottom-up and top-down feature propagation pathways to improve the network model.

2)	We propose both coarse and fine-grained refinement networks in the learning process. The coarse refinement branch provides initial prediction to bootstrap saliency estimation, while the fine-grained refinement branch serves as a dedicated detector to refine the previous prediction and locate the salient object in fine detail from background regions.

3)	A novel attention mechanism is proposed to make full use of cross-modal visual cues for accurate detection results. The proposed technique can efficiently mine correlation between two modals, and reduce information loss of the salient model.

4)	We present extensive qualitative and quantitative experiments to validate the effectiveness of our proposed method. The evaluation results demonstrate our proposed method performs favorably against state-of-the-art methods.

Our paper is organized in the following structure. Section 2 gives a short review of the recent studies which are related to our work. Section 3 presents the detailed architecture of our proposed method. Experimental evaluations as well as ablation studies are given in Section 4. Section 5 draws a conclusion of our work. We hope our research can give some insights for future studies.

\section{Related work}

\subsection{Salient Object Detection}

Salient Object Detection methods can be divided into hand-crafted method and deep learning based method. Early detection methods rely on contrast prior to extract saliency maps. For example, 
\cite{20itti1998model} proposes a model to generate saliency maps from color contrasts, intensity contrasts and orientation contrasts. 
\cite{21klein2011center} computes image intensities, colors, and orientations with multiple scales, and the center-surround contrast based on information theory is devised to mine various feature cues. 
\cite{22cheng2014global} follow a bottom-up process which simultaneously considers both global contrast and spatial weighted coherence scores to generate accurate results. 
\cite{23perazzi2012saliency} incorporates predefined contrast prior which focus on uniqueness and spatial distribution of visual image. 
\cite{24liu2010learning} formulates a unified approach which considers multi-scale contrast, center-surround histogram, and color spatial distribution to detect salient object.

In contrast to aforementioned methods, Later approaches leverage current advanced of deep learning to predict saliency regions. 
\cite{25li2016visual} constructs deep contrast model which separates foreground objects from multi-scale discriminative features and handcrafted low-level features. 
\cite{13li2016deep} develops an end-to-end deep contrast network, which consists of a CRF model leveraging both pixelwise and segment-wise features to produce saliency map. 
\cite{26kruthiventi2016saliency} incorporates object level semantics and global context to enforce the network to mimic eye fixations. In order to improve the robustness of salient detection, 
\cite{27zhang2017learning} introduces uncertain convolutional features as well as reformulated dropout layer into neural network. Inspired by Mumford-Shah function 
\cite{28pock2009algorithm}, 
\cite{29luo2017non} formulates a loss function along with global and local grid-structure to separate conspicuous objects. 
\cite{30zhang2018progressive} introduces recurrent model into salient object detection, which incorporates multi-level constext and multi-path recurrent feedback reinforce discriminative learning process. To address on boundary estimation, 
\cite{31qin2019basnet} develops predict to refine approach, which consists of U-Net based encoder-decoder network as well as a refinement neural network. More recent method 
\cite{10pang2020multi} utilizes self and adjacent interactions to exploit multi-level and multi-scale features.

\subsection{RGB-D Salient Object Detection}

RGB-D salient object detection can be roughly classified into early fusion, middle fusion and late fusion. The first direction, regards depth image as one channel, and applies simple transformation to fuse the RGB channel. 
\cite{16peng2014rgbd} incorporates per pixel multiplication for RGB image and depth map in the early stage. A contrast based analysis is employed to explore visual saliency.
\cite{48fan2014salient}
 leverages weights between regional segmentation of stereoscopic image and SIFT based depth contrast to generage saliency map.
 \cite{32chen2021rgb} regards depth as one channel of RGB image, and develops 3D convolution based neural network to leverage cross-modal correlation. 
\cite{33liu2019salient} uses four channel RGB-D input, both single stream and depth recurrent neural network are applied to each level features. The final result is generated by fusing multiple level features.

Late fusion method introduces post processing step to combine separate RGB and depth processing streams. 
\cite{18cheng2014depth} computes color contrast, depth contrast and spatial bias, and generates saliency map by later combining various saliency cues. 
\cite{34desingh2013depth} takes into account object structure information to compute depth saliency map and 2D saliency map. The aforementioned saliency results are fused by SVM regressor. 
\cite{35zhu2017innovative} begins by generating color based saliency map and center-dark channel map, and then fuses two feature maps to formulate final result.

Middle fusion method aims to integrate cross-modal features in the middle stage. 
\cite{5piao2019depth} incorporates refinement blocks with residual connections to exploit complementary information between RGB and depth channels. In order to sufficiently incorporate complementary part of cross modal features,
 \cite{7chen2018progressively} presents a cascaded complementary-aware fusion with cross-modal residual function to effectively mine visual cues from different channels. 
\cite{9chen2019multi} focuses on multi-layer cross-modal interactions in the middle stage, which leads to flexible fusion stream and multi-folds integration. 
\cite{36li2020cross} develops various interaction modules (CMW-L, CMW-M and CMW-H) which fuse information between RGB and depth channels. In 
\cite{37zhu2019pdnet}, the proposed network consists of master network, which is based on convolution-deconvolution pipeline, and sub-network, which focuses on depth map processing and assisting the master network in the middle stage learning. 
\cite{38chen2019three} introduces three stream networks where cross-modal cross-level fusion is leveraged to explore attention based visual cues. 
\cite{39zhai2021bifurcated} introduces teacher based guidance to refine detector and leverages multi-level multi-modal fusion to improve detection accuracy. 
\cite{40sun2021deep} extracts depth map into regional features and leverages multi-modal fusion scheme with depth-aware attention module for salient object detection.

As our method iteratively combines cross-modal visual cues in attention based manner, our method falls into the middle fusion class.

\subsection{Multi-scale Learning}

Multi-scale learning plays an important role in deep learning to augment feature representation.  
\cite{41zhang2017amulet} focuses on extracting multi-level features at different resolutions. Edge-aware feature maps are integrated into well-defined aggregation process. 
\cite{42zhang2018bi} leverages bi-directional message passing mechanism to aggregate from various level features. However, gating mechanism leads to degraded message integration from other layers, which is sub-optimal in multi-scale modeling. The atrous spatial pyramid pooling (ASPP) 
\cite{43chen2017deeplab} is naturally to model multi-scale information. Different filters with various sampling rates are beneficial to combine multi-scale context. Nevertheless, ASPP layer is incapable to mine relations between various layers with large scale variations. To address on these issues, 
\cite{9chen2019multi} proposes multi-scale multi-path fusion to reinforce multi-scale learning in RGB-D salient object detection. However, without deep supervision of multiple level features, these approaches lead to sub-optimal in feature propagation. In this paper, we introduce iterative refinement with multiple level supervision to reinforce learning process.

\section{Multi-scale Iterative Refinement Neural Network}

In this section, we first formulate the overall scheme of our proposed method in $Sec. 3.1$, and then introduce the $K - fold$ coarse refinement in  $Sec. 3.2$. The top feature $e_6 $ is fed into fine-grained refinement block to learn enhanced saliency map. Please see details in $Sec. 3.3$. The formulation of our iterative deep supervision is presented in $Sec. 3.4$. The notations used throughout the paper are shown in Table 1.
\begin{figure}[H]
	\centering
	\includegraphics[scale=0.7]{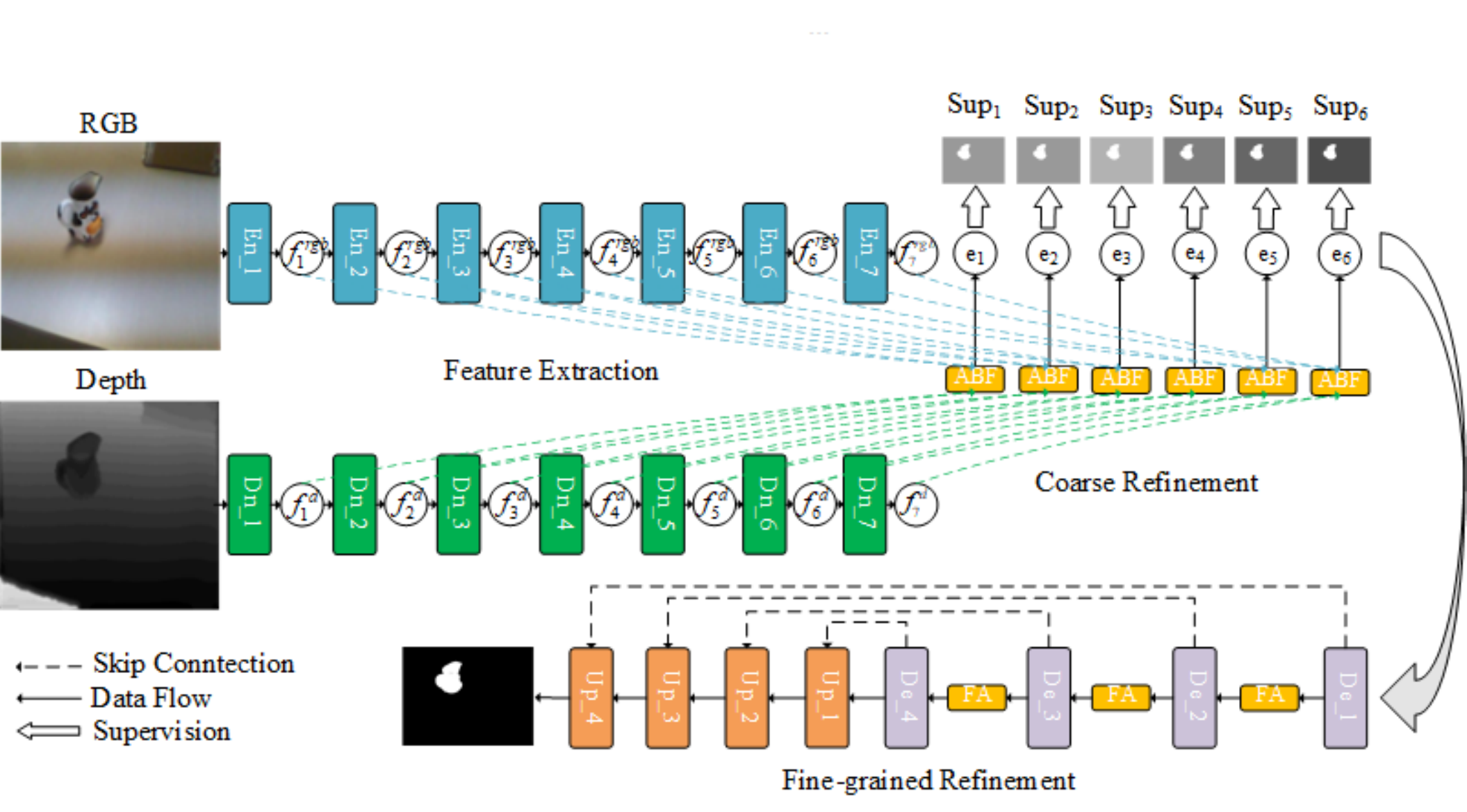}
	\caption{ The Architecture of our proposed MSIRN, where cross-modal features are fed into ABF module for iterative refinement. The proposed method contains 3 stages: feature extraction $En_1  - En_7 ,Dn_1  - Dn_7 $, coarse refinement $ABF$ and fine-grained refinement $De_1  - De_4 ,Up_1  - Up_4 ,FA$ . The cross-modal features $f_1^{rgb}  - f_7^{rgb} ,f_1^d  - f_7^d $ are generated by two-stream network. To obtain coarse saliency map, we integrate the visual cues in a top-down fashion by multiple ABF modules. We take the top feature $e_6 $ to incorporate fine-grained refinement.}
	\label{fig2}
\end{figure}

\subsection{Architecture Overview}

Our network can not only leverage cross-modal visual cues, but also deal with multi-scale level features. In this paper, we consider the learning problem of RGB-D salient object detection task as iterative $k - fold$  coarse refinement as well as a fine-grained refinement, which is denoted as $R_c  = \left\{ {R_c^1 , \cdots ,R_c^k } \right\}$
 and $R_f $. As depicted in Fig. 2, our model contains several bottom-up encoder blocks ($Ens,Dns,Des$), upsample blocks ($Ups$), attention based fusion modules ($ABFs$) and feature aggregation modules ($FAs$).

\begin{table}[H]
  \centering
  \caption{Nomenclature}
    \begin{tabular}{cc}
    \toprule
    \multicolumn{1}{p{2.585em}}{Symbol} & Description \\
    \midrule
$f_i^{rgb}$         & The i-th level RGB feature \\
$f_i^d $          & The i-th level depth feature \\
$e_i $         & The i-th output feature of coarse refinement branch \\
$R_c^i$          & The i-th coarse refinement block \\
$S_i^c $          & The i-th output saliency map of coarse refinement branch \\
${\cal F}_i $
          & The i-th level encoding feature map of fine-grained refinement branch \\
${\cal O}_i $
          & The i-th level decoded feature map of fine-grained refinement branch \\
$G$
          & The ground truth saliency map \\
$S^f $
          & The predicted saliency map of fine-grained refinement branch \\
$U\left(  \cdot  \right)$
          & The upsample operation \\
$ \otimes $
          & Element-wise multiplication broadcasted along feature planes \\
$Cat\left(  \cdot  \right)$   
          & Channel-wise concatenation operation \\
$GP\left(  \cdot  \right)$
          & Global average pooling function \\
$\mathcal{L}\left(  \cdot  \right)$
          & The binary cross-entropy loss \\
    \bottomrule
    \end{tabular}%
  \label{tab:addlabel}%
\end{table}%

\textbf{Feature extracting.}The RGB and depth information are encoded by two stream networks. Inspired by ResNet-34 , we employ 7 layers of input convolution for both RGB and depth streams. We adopt the first 4 layers from ResNet, and add 3 additional layers to improve the overall performance. With deeper network, the high-level feature maps (e.g.,$f_6^{rgb} $) are able to extract more contextual information. Different from ResNet, we employ $3 \times 3$ convolutional layer instead of $7 \times 7$ convolutional filter for the first layer. In addition, we remove all max pooling operation, which means our scale size is $1/2$ instead of $1/4$. The $i$-th layer feature $f_i^{rgb} $ of RGB stream and $f_i^d $ of depth stream as well as higher layer features $f_{i + 1}^{rgb} $ and $f_{i + 1}^d $ are fed into the $i$-th coarse refinement module ABF for cross-modal modeling.

\subsection{K-fold coarse refinement}

Our goal is to explore potential multi-level features in multiple stages which bridge the semantic gaps between different scales. In contrast to early fusion or late fusion methods , we exploit multi-modal information in iterative refinement process, rather than one or two steps. Our refinement process is divided into coarse and fine-grained refinement. The coarse refinement process adopts a top-down fashion, which progressively fuse cross-scale multi-modal context to produce more efficient estimation features. The coarse term here has two characteristics. One is blurry, the other is the uncertainty of saliency probabilities. The $f_i^{rgb} $ and $f_i^{depth} $ from lower layer have less semantic features but beneficial to locate object’s local context. Thus, we feed $f_i^{rgb} $ and $f_i^{depth} $ as well as higher level features $f_{i + 1}^{rgb} $ and $f_{i + 1}^{depth} $ to ABF module to predict saliency map, that is

\begin{equation}
e_i  = Cat\left( {U\left( {ASPP\left( {Conv\left( {Cat\left( {f_i^{rgb} ,f_i^{depth} } \right)} \right)} \right)} \right) + Att\left( {Cat\left( {f_{i + 1}^{rgb} ,f_{i + 1}^{depth} } \right)} \right)} \right),i \in 1,2,3,4,5,6
\label{1}
\end{equation}

where $Cat\left(  \cdot  \right)$ is the channel-wise concatenation operation,$Att$ is the attention based cross-modal fusion module,$U\left(  \cdot  \right)$ is the upsample operation. Based on the limit of receptive field of convolutional filter, we adopt atrous spatial pyramid pooling (ASPP) to enrich the semantic feature with larger receptive field. Specifically, the ASPP layer aggregates scaled feature maps with various dilated rates, e.g. 1, 6, 12, 18. In order to integrate adjacent $i$-th and  $\left( {i + 1} \right)$-th layer, we use bilinear upsample method to scale the $i$-th feature map $R^{H/2 \times W/2} $ to $R^{H \times W} $. To efficiently improve the discriminative representation of cross-modal features, we leverage channel-wise and spatial attention to fuse multi-modal features. As illustrated in Fig. 3, spatial attention and channel-wise attention are employed separately. Given concatenated layer feature $f = \left[ {f_1 ,f_2 ,f_3 ,f_4 ,f_5 ,f_6 } \right]$ generated by encoding stage, we formulate the channel-wise attention as

\begin{equation}
Att_{channel} \left( i \right) = a_{channel} \left( i \right) \otimes f_i ,i \in 1,2,3,4,5,6
\label{2}
\end{equation}

attentive channel vector $a_{channel} \left( i \right)$ is obtained by

\begin{equation}
a_{channel} \left( i \right) = M\left( {MP\left( {f_i } \right)} \right),i \in 1,2,3,4,5,6
\label{3}
\end{equation}

where $M\left(  \cdot  \right)$ is the convolution operation followed by batch normalization and ReLU activation,$MP\left(  \cdot  \right)$ is the max pooling layer,$ \otimes $ denotes broadcast multiplication. 

On the other hand, the spatial attention is defined as

\begin{equation}
Att_{spatial} \left( i \right) = a_{spatial} \left( i \right) \times f_i ,i \in 1,2,3,4,5,6
\label{4}
\end{equation}

where $a_{spatial} \left( i \right)$ is defined as

\begin{equation}
a_{spatial} \left( i \right) = Conv\left( {GP\left( {f_i } \right)} \right),i \in 1,2,3,4,5,6
\label{5}
\end{equation}

where $GP\left(  \cdot  \right)$ is global average pooling function.

\begin{figure}[H]
	\centering
	\includegraphics[scale=0.7]{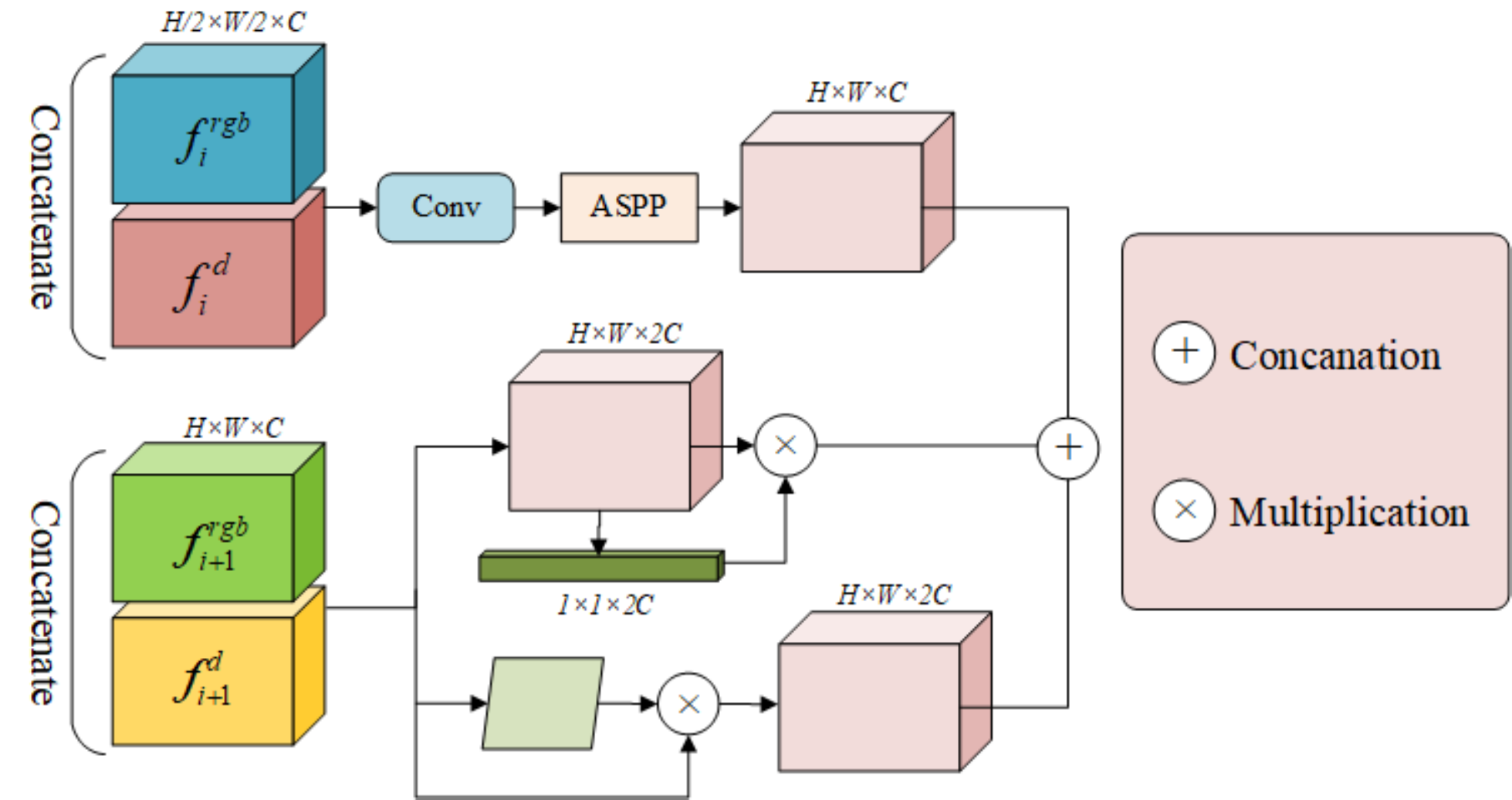}
	\caption{Details of our attention based fusion (ABF) module, which integrates features from $\left( {i + 1} \right)$-th and $i$-th layer, low-level features are upsampled to enhance local features, while high-level cross-modal features are explored with both channel-wise and spatial attention.}
	\label{fig3}
\end{figure}

In order to enhance cross-modal visual cues, we concatenate attention based multi-modal features and lower level features to excavate multi-scale learning. To achieve the output saliency map, each concatenated feature $e_i \left( {i = 1,2,...,6} \right)$ is fed to a $3 \times 3$ convolution layer followed by a upsampling function, and activated by a sigmoid function. In this way, six multi-level predictions $S_i^c \left( {i = 1,2,...,6} \right)$ with the same size but different context are generated.

\subsection{Fine-grained refinement}

Due to top-down based feature aggregation, the high-level feature degrades in repeated up-sampling operations, which leads to blurred or obscured saliency maps. As a result, it is insufficient to predict saliency objects by only leveraging coarse refinement.

To accurately extract the most conspicuous object and discriminate salient objects from background regions, we propose to mitigate the ambiguity of saliency map by fine-grained refinement with additional guidance. Inspired by U-Net 
\cite{66ronneberger2015u}, we design our refinement network in an encoder-decoder fashion. However, U-Net method, which directly incorporates convolution and deconvolution operation, is defective due to the limited receptive fields and inefficient feature mining. With this in mind, we exploit dilated convolution and introduce feature aggregation module (FA) to produce enhanced saliency features.

\begin{figure}[H]
	\centering
	\includegraphics[scale=0.7]{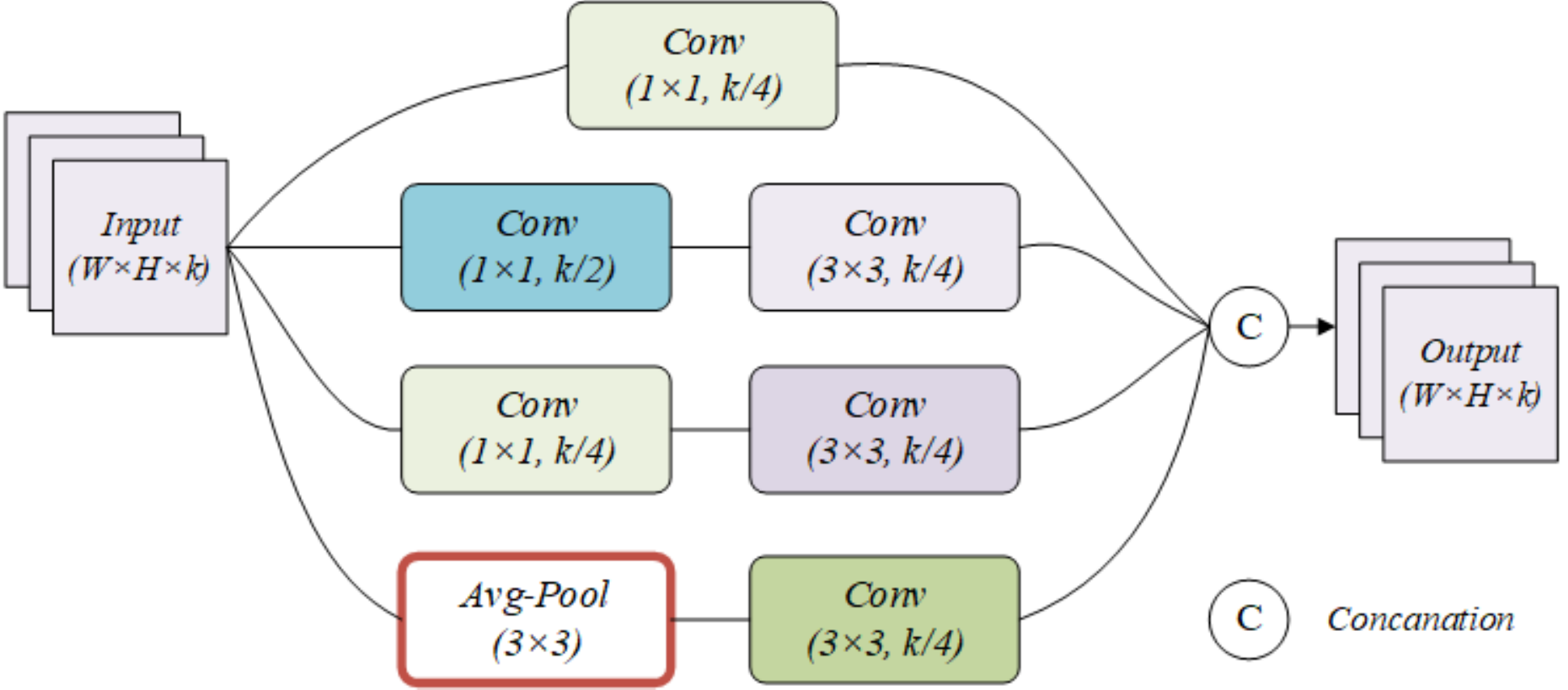}
	\caption{Details of our feature aggregation (FA) module, which integrates features with different receptive fields. Three convolutional layers and an average pooling layer are leveraged to enlarge receptive fields.}
	\label{fig4}
\end{figure}

As shown in Fig. 2, cross-modal feature cues $e_6 $ with abundant semantic information is further leveraged to learn a fine-grained saliency map. Specifically, in order to enhance local features and mitigate blurry and uncertainty of saliency map, we impose three feature aggregation modules with diverse receptive fields, several encoding blocks $De\_1 - De\_4$ and symmetrical decoding blocks $Up\_1 - Up\_4$ with residual connections. We employ convolutional layer with different dilation rates and max pooling operation in each level. In this way, feature map with enhanced contexts is generated. Compared with canonical convolution layer, dilated convolution operation is beneficial to enlarge the size of receptive field without increasing computational workload. At the same time, in order to obtain the enhanced version of feature representation, three feature aggregation layers are imposed after each encoding layer in the refinement stream. Fig. 4 illustrates feature aggregation blocks with various receptive fields. The output feature maps of canonical convolution layer limit local features because simple convolutional filters of $1 \times 1$ or $3 \times 3$ are insufficient to extract local information. To efficiently exploit salient information of feature maps from various layers, the most direct idea is to use various convolutional kernels as well as pooling operations to enhance feature representation. Inspired by inception block , we propose to extract both pooling based and convolutional based features and aggregate in late fusion step. Encoded feature maps are further fed to decoding stage. However, conducting multiple up-sampling layer in a top-down manner in the decoding stage suffers from degradation of detailed information caused by direct up-sampling. To this end, we employ residual connection to further improve the contextual information of the decoding stage, where $\mathcal{F}_i $ responds to the encoding feature map of level $i$, and $\mathcal{O}_i $ denotes the up-sampled feature map of level $i$. Formally, the skip-connected decoding blocks can be defined as:

\begin{equation}
\mathcal{O}_i  = U\left( {Conv\left( {Cat\left( {\mathcal{O}_i ,\mathcal{F}_i } \right)} \right)} \right),i \in 1,2,3,4
\label{6}
\end{equation}

\subsection{Deep supervision}

\begin{figure}[H]
	\centering
	\includegraphics[width=\textwidth]{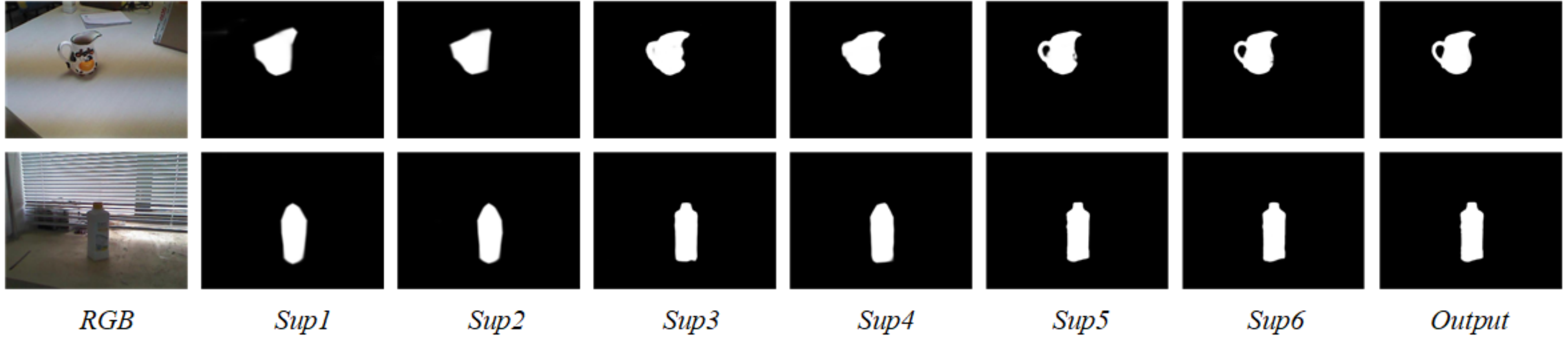}
	\caption{Illustration of our iterative refinement saliency maps with deep supervision.}
	\label{fig5}
\end{figure}

To obtain high quality salient maps and clear object regions, we propose to incorporate iterative supervision to refine object and progressively fuse cross-modal visual cues. We formulate the loss as the summation of K-fold coarse refinement loss as well as a fine-grained refinement loss.

\begin{equation}
\mathcal{L}_{total}  = \mathcal{L}_f \left( {S^f ,G} \right) + \lambda \mathop \sum \limits_{i = 1}^K \mathcal{L}_c \left( {S_i^c ,G} \right)
\label{7}
\end{equation}

where $\lambda $ balances the importance between coarse refinement and fine-grained refinement,$Sup1$,$Sup2$,
$Sup3$,$Sup4$,$Sup5$,$Sup6$ in Fig. 2 refer to K-fold refinement branch,$G$ is the supervision from the ground truth saliency map,$S^f $ is the output from fine-grained refinement branch,$S_i^c $ is the output from coarse refinement branch. BCE  loss is naturally to model the classification or segmentation. For each term, we adopt the standard binary cross-entropy to calculate the loss:

\begin{equation}
\mathcal{L}\left( {S,G} \right) = GlogS + \left( {1 - G} \right)log\left( {1 - S} \right)
\label{8}
\end{equation}

where $G \in 0,1$ is the ground truth label of each pixel of saliency map, while $S$ is the predicted probability of saliency map.
The goal of training stage is to minimize the overall loss of Eq. (7), while in the testing stage, we choose the final output $S^f $ as our final predicted saliency map.

\section{Experimental Design and Comparative Results}

In this section, we conduct extensive experiments to validate the effectiveness of our proposed method. We will illustrate the merits of our proposed MSIRN method from various aspects. We first compare the performance of our MSIRN network with other state-of-the-arts salient object detection methods. We use iterative refinement to progressively refine the salient maps reinforced by coarse-to-fine supervision. It is worth mentioning that the low-level refinement results still perform well in the salient object detection task. In addition to quantitative experiments, we also conduct qualitative experiments to analyze the performance of our proposed method. In our next set of experiments, we intend to verify the contribution of different components in MSIRN. As can be seen from the following experimental results, our devised MSIRN network achieves great advantages on the majority of data sets and plays favorably against state-of-art methods. Finally, to highlight the uncertainty of the saliency map, we check the failure cases of the proposed model. 

In the following sections, we first give the implementation details and training setup in Section 4.1. In Section 4.2, we introduce the public datasets used in our experiments. The comparison algorithms and evaluation metrics are elaborated in Section 4.3 and Section 4.4. Qualitative and quantitative comparisons are given in Section 4.5, followed by ablation study in Section 4.6.

\subsection{Implementation Details}

Our framework is built on top of PyTorch framework. All experiments are conducted on a single NVIDIA 1080ti GPU. SGD optimizer is ultilized with learning rate of 0.0001. In all experiments, we implement 7 layers of feature extraction module and deploy coarse refinement branch with refinement steps $K = 6$. We employ 4 layers of encoder and 4 symmetrical decoder layers in fine-grained refinement branch. The ASPP layer aggregates scaled feature maps with dilated rates of  1, 6, 12, 18. We adopt bilinear interpolation in upsampling process. In order to prevent model degradation, skip connections are added to the input of the hidden layers of decoder. For loss function, the balance variable $\lambda$ dynamically weights between coarse refinement and fine-grained refinement learning process, the initial balance variable is set to 0.99. In addition, we decay the learning rate by 0.1 for every 60 epochs. Following 
\cite{47de2005tutorial}, we exploit 1485 images and 700 images from \emph{NJU2K} 
\cite{48fan2014salient} and \emph{NLPR} 
\cite{16peng2014rgbd} datasets to train our network. We set the total learning epochs to 300 in total, and it takes about 7 hours to optimize the learning parameters. The batch size is set to 8. The initial learning parameters are loaded from ResNet-34 model pretrained on ImageNet dataset. For data preprocessing, both RGB and depth image are resized to 224×224. During testing, the output from model is further resized to the original resolution.

\subsection{Dataset Description}

Extensive experiments are conducted on seven commonly used datasets in salient object detection domain, including \emph{NJU2K},\emph{NLPR},\emph{STERE},\emph{DES},\emph{LFSD},\emph{SSD} and \emph{SIP}.

\textbf{\emph{NJU2K}}
 \cite{48fan2014salient} is a large scale dataset which contains 1,985 images and depth pairs. This dataset is popular for a wide range of RGB-D dataset. All images are collected by photographers from Internet or movies.

\textbf{\emph{NLPR}}
 \cite{39zhai2021bifurcated} is a medium size dataset formed by 1,000 image pairs. The depth images are captured by a Miscrosoft Kinect device. The resolution of paired image is 640×480. This dataset contains wide range of scenes including indoors, offices, streets and shops. 

\textbf{\emph{STERE}}
 \cite{49niu2012leveraging} is the first stereoscopic dataset which consists of 1,000 image pairs captured by stereoscopic images. Notably there are two versions of this dataset. One contains 979 images, while the other contains 1,000 images. Here, we adopt the 1,000 version to conduct experiments.

\textbf{DES}
 \cite{18cheng2014depth} is a small dataset used in our experiments, which contains 135 image pairs. The dataset is captured by Miscrosoft Kinect device in indoor scenes with resolution of 640×480. 

\textbf{\emph{LFSD}}
 \cite{15li2014saliency} is the first collected dataset which leverages Lytro light field camera. The dataset contains 100 image pairs, where 60 image pairs are collected from indoor scenes, and 40 image pairs are captured from outdoor scenes. The resolution of image pairs is 640×480.

\textbf{\emph{SSD}} 
 \cite{51zhu2017three} is collected from three movies and captured by stereo cameras. The dataset contains 80 images from both indoor and outdoor scenes. The images have high resolution of 960×1080.

\textbf{\emph{SIP}}
 \cite{50fan2020rethinking} is a collection of 1,000 image pairs from mobile phones. The photo device leverages dual camera, which has high resolution of 992×744. The research interest of this dataset is real world salient person.

\subsection{Comparison Algorithms}

The proposed MSIRN is compared with 15 baseline methods. The compared methods can be divided into 2 categories: traditional methods and deep neural network based methods. We include 5 traditional methods in our analysis: Anisotropic Center-Surround Difference (ACSD) 
\cite{52ju2014depth}, which takes the global depth information into consideration and introduces anisotropic center-surround difference to model depth images; Local Background Enclosure (LBE) 
\cite{19feng2016local}, which incorporates distance between object and background regions from depth map to enhance feature representation; Depth Confidence and Multiple Cues (DCMC) 
\cite{53cong2016saliency}, which constructs graph from depth map and utilizes multiple cues contrast; Multiscale Discriminative Saliency Fusion (MDSF) 
\cite{54song2017depth}, which attempts to learn different levels of contrasts to perform saliency prediction; Saliency evolution (SE) 
\cite{55guo2016salient} aims to generate initial saliency map, and then propagates the initial saliency to the final detection result. In addition, we also include 9 deep learning based methods, such as Deep Fusion (DF) 
\cite{64qu2017rgbd}, which incorporates Laplacian framework for feature propagation; Adaptive Fusion (AFNet) 
\cite{56wang2019adaptive}, which dynamically fuses between RGB and depth predictions; Cross-View Transfer and Multiview Fusion (CTMF) 
\cite{57han2017cnns}, which proposes to leverage cross-view transfer and multiview fusion in salient detection; Multi-scale Multi-path Cross-modal interactions (MMCI) 
\cite{9chen2019multi} introduces two-stream network with multi-path and multi-scale fusion; Progressively Complementarity-Aware Fusion (PCF) 
\cite{7chen2018progressively}, which introduces complementarity-aware fusion in multiple levels in detection stream; Three-stream Attention-aware Network (TANet) 
\cite{38chen2019three} employs three stream attention based neural network to learn salient objects; Contrast Prior and Fluid Pyramid (CPFP) 
\cite{6zhao2019contrast}, which leverages contrast prior to better model multi-scale cross-modal features; Depth-induced Multi-scale Recurrent Attention Network (DMRA) 
\cite{5piao2019depth}, which devises a recurrent attention module to boost detection performance; D3Net 
\cite{50fan2020rethinking}, which employs three-stream feature learning module to facilitate the learning process; Multi-level Cross-modal Interaction Network (MCI-Net) 
\cite{58huang2021multi}, which exploits dedicated fusion module to facilitate multi-level cross-modal fusion.

\subsection{Evaluation Metrics}

To compare the performance of various methods, we adopt widely-used metrics to analyze the performance of different salient detection methods. The evaluation metrics include Precision-Recall curve (PR), F-measure, Mean Absolute Error (MAE), S-measure ($S_\alpha  $) and E-measure ($E_\xi  $).

\textbf{\emph{PR curve}} 
\cite{59powers2011evaluation} is a commonly used evaluation metric for salient object detection. PR curve evaluates precision-recall (PR) pairs with a series of binary thresholds from 0 to 255. Formally, we use the following equations to calculate the precision (P) and recall (R):

\begin{equation}
P = \frac{{\left| {S' \cap G} \right|}}
{{\left| {S'} \right|}}
\label{9}
\end{equation}

\begin{equation}
R = \frac{{\left| {S' \cap G} \right|}}
{{\left| G \right|}}
\label{10}
\end{equation}

where $S'$ is the binary map with given threshold,$S$ is predicted saliency map,$G$ is ground truth saliency map.

\textbf{\emph{F-measure}}
 \cite{60achanta2009frequency} is a metric considering both precision and recall. F-meature can be formulated as

\begin{equation}
F_\beta ^i  = \frac{{\left( {1 + \beta ^2 } \right)P^i  \times R^i }}
{{\beta ^2  \times P^i  + R^i }}
\label{11}
\end{equation}

where $P^i$ and $R^i$ are precision and recall for specific threshold $i\left( {i \in \left\{ {1,2,...,255} \right\}} \right)$.$\beta$ is the weight variable between $P^i$ and $R^i$. We follow 
\cite{60achanta2009frequency} and set $\beta ^2  = 0.3$. To make comprehensive analysis, we also include maximum F-measure, adaptive F-measure and mean F-measure in our experiments.

\textbf{\emph{Mean Absolute Error}}
 \cite{61perazzi2012saliency} evaluates the mean absolute error between predicted saliency maps and ground truth binary maps. It is defined as

\begin{equation}
M = \frac{1}{N}\left| {S - G} \right|
\label{12}
\end{equation}

where $N$ is the total number of pixels.

\textbf{\emph{S-measure}}
 \cite{62fan2017structure} is an advanced metric, which takes both regional and object structural similarity into account. Specifically, S-measure metric can be formulated as

\begin{equation}
S_\alpha   = \alpha  \times S_o  + \left( {1 - \alpha } \right) \times S_r 
\label{13}
\end{equation}

where $\alpha  \in \left[ {0,1} \right]$ is a weighting variable between object structural similarity $S_o $ and regional similarity $S_r $. Here we set $\alpha  = 0.5$.

\textbf{\emph{E-measure}}
 \cite{63fan2018enhanced} is a recent proposed metric, which leverages both global and local pixel-level analysis. The metric is formulated as

\begin{equation}
E_\xi   = \frac{1}
{{w \times h}}\sum\limits_{x = 1}^w {\sum\limits_{y = 1}^h {\xi \left( {x,y} \right)} }
\label{14}
\end{equation}

where $w$ and $h$ are width and height of the input image,$\xi $ is the alignment matrix. For comprehensive comparison, we compute max, adaptive and mean E-measure in all our experiments.

\subsection{Qualitative and Quantitative Comparisons}

We follow the same data split protocal in D3Net 
\cite{50fan2020rethinking} and generate saliency maps using our proposed MSIRN model. To analyse the detection performance, we compare our method with 10 deep learning based methods(e.g., TIP’17 DF 
\cite{64qu2017rgbd}, arXiv’19 AFNet 
\cite{56wang2019adaptive}, TOC’18 CTMF 
\cite{57han2017cnns}, PR’19 MMCI 
\cite{9chen2019multi}, CVPR’18 PCF 
\cite{7chen2018progressively}, TIP’19 TANet 
\cite{38chen2019three}, CVPR’19 CPFP 
\cite{6zhao2019contrast}, ICCV’19 DMRA 
\cite{5piao2019depth}, TNNLS’20 D3Net 
\cite{50fan2020rethinking}, Neurocomputing’21 MCI-Net 
\cite{58huang2021multi}) and 5 tranditional methods (ICIP’14 ACSD, CVPR’16 LBE, SPL’16 DCMC, TIP’17 MDSF, ICME’16 SE). Table 2 shows that our proposed method MSIRN outperforms other methods across seven datasets. More specifically, PR’19 MMCI 
\cite{9chen2019multi}, CVPR’18 PCF 
\cite{7chen2018progressively}, ICCV’19 DMRA 
\cite{5piao2019depth} and Neurocomputing’21 MCI-Net 
\cite{58huang2021multi} also integrate cross-modal multi-level features to facilitate the learning process in middle stage fashion. However, our method leverages iterative refinement learning process, which involves coarse-to-fine saliency refinements and alleviates the uncertainty in prediction probabilities.

\textbf{Performance of traditional methods.} Comparing the performance in Table 2, DCMC 
\cite{53cong2016saliency}, MDSF 
\cite{54song2017depth} and SE 
\cite{55guo2016salient} are the best 3 models. All of these methods evaluate regional contrast based on RGB appearance or depth spatial information. However, MDSF 
\cite{54song2017depth} exploits bootstrap learning and multi-scale fusion, which leads to better results.

\textbf{Performance of deep learning methods.} Among deep learning based methods, our MSIRN, MCI-Net 
\cite{58huang2021multi} and D3Net 
\cite{50fan2020rethinking} are the top leading methods. These methods conduct explicit feature level deep fusion in multiple steps, which effectively take advantage of complementary information in multi-modal features.

\textbf{Deep learing methods} \emph{vs} \textbf{Tradional methods.} As shown in Table 2, deep learning based methods outperform traditional methods in a large margin. However, on \emph{SIP} dataset, tradional method ACSD 
\cite{52ju2014depth} outperforms 3 deep learning based methods DF 
\cite{64qu2017rgbd}, AFNet 
\cite{56wang2019adaptive} and CTMF 
\cite{57han2017cnns} in terms of S-measure ($S_\alpha  $), F-measure (max $F_\beta  $, mean  $F_\beta  $, adp $F_\beta  $) and E-measure (max $E_\xi  $, mean $E_\xi  $, adp $E_\xi  $).

Benefitting from more efficient cross-modal fusion, our method performs favorable against state-of-the-art methods. Specifically, as presented in Table 2, our method achieves 1.2$\%$, 1.5$\%$, 0.3$\%$, 0.4$\%$, 1.4$\%$, 1.5$\%$ and 1.2$\%$ performance imrpovements in terms of $S_\alpha  $ among 15 competing methods on \emph{NJU2K}, \emph{LFSD}, \emph{STERE}, \emph{DES}, \emph{NLPR}, \emph{SSD} and \emph{SIP} datasets respectively. In addition, the proposed method outperforms other methods by nearly 2.0$\%$ performance boost in terms of F-measure (max $F_\beta  $, mean  $F_\beta  $, adp $F_\beta  $), E-measure (max $E_\xi  $, mean $E_\xi  $, adp $E_\xi  $), and promising decrease in terms of mean absolute error (MAE). Moreover, competing with more recent MCI-Net 
\cite{58huang2021multi}, we observe more favorable performance, which shows the effectiveness of our proposed iterative refinement and cross-modal fusion. Notably, our method produces nearly 1.2$\%$ improvement on the challenging real-world \emph{SIP} dataset with complex scenes.

\begin{table}[H]
\centering
\caption{Quantitative comparisons between our proposed MSIRN and other 15 methods: including traditional methods and deep learning based methods. Extensive experiments are conducted on 7 datasets using S-measure ($S_\alpha$), max F-measure (max $F_\beta$), mean F-measure (mean $F_\beta$), adaptive F-measure (adaptive $F_\beta$), max E-measure (max $E_\xi$), adaptive E-measure (adaptive $E_\xi$) and mean absolute error (MAE). ↑ and ↓ denote that the higher or lower is better.}
\resizebox{\textwidth}{!}
{
\begin{tabular}{ccccccc|cccc}
\hline
\hline
 & &ACSD&LBE&DCMC&MDSF&SE&DF&AFNet&CTMF&MMCI\\
 &Metric&ICIP14&CVPR16&SPL16&TIP17&ICME16&TIP17&arXiv19&TOC18&PR19\\
 & &[52]&[19]&[53]&[54]&[55]&[64]&[56]&[57]&[9]\\
\hline
\hline
\multirow{8}*{\rotatebox{90}{\emph{NJU2K} 
\cite{48fan2014salient}}} &$S_\alpha$↑      & 0.699    & 0.695    & 0.686    & 0.748    & 0.664    & 0.763    & 0.772    & 0.849    & 0.858   \\ 
 &max $F_\beta$↑   & 0.711    & 0.748    & 0.715    & 0.775    & 0.748    & 0.804    & 0.775    & 0.845    & 0.852   \\
 &mean $F_\beta$↑  & 0.512    & 0.606    & 0.556    & 0.628    & 0.583    & 0.65     & 0.764    & 0.779    & 0.793   \\
 &adp $F_\beta$↑   & 0.696    & 0.74     & 0.717    & 0.757    & 0.734    & 0.784    & 0.768    & 0.788    & 0.812   \\
 &max $E_\xi$↑   & 0.803    & 0.803    & 0.799    & 0.838    & 0.813    & 0.864    & 0.853    & 0.913    & 0.915   \\
 &mean $E_\xi$↑  & 0.593    & 0.655    & 0.619    & 0.677    & 0.624    & 0.696    & 0.826    & 0.846    & 0.851   \\
 &adp $E_\xi$↑   & 0.786    & 0.791    & 0.791    & 0.812    & 0.772    & 0.835    & 0.846    & 0.864    & 0.878   \\
 &$\mathcal{M}$↓      & 0.202    & 0.153    & 0.172    & 0.157    & 0.169    & 0.141    & 0.100    & 0.085    & 0.079 \\ 
\hline
\multirow{8}*{\rotatebox{90}{\emph{LFSD} 
\cite{15li2014saliency}}} &$S_\alpha$↑&0.862 & 0.736 & 0.753 & 0.700 & 0.698 & 0.791 & 0.738 & 0.826 & 0.787 \\
 &max $F_\beta$↑   &0.863 & 0.726 & 0.817 & 0.783 & 0.791 & 0.817 & 0.744 & 0.791 & 0.771 \\
 &mean $F_\beta$↑  &0.838 & 0.611 & 0.655 & 0.521 & 0.640 & 0.679 & 0.735 & 0.756 & 0.722 \\
 &adp $F_\beta$↑   &0.851 & 0.708 & 0.816 & 0.799 & 0.778 & 0.806 & 0.742 & 0.782 & 0.779 \\
 &max $E_\xi$↑   &0.898 & 0.804 & 0.856 & 0.826 & 0.840 & 0.865 & 0.815 & 0.865 & 0.839 \\
 &mean $E_\xi$↑  &0.873 & 0.670 & 0.683 & 0.588 & 0.653 & 0.726 & 0.796 & 0.810 & 0.775 \\
 &adp $E_\xi$↑   &0.885 & 0.770 & 0.842 & 0.817 & 0.784 & 0.844 & 0.810 & 0.851 & 0.840 \\
 &$\mathcal{M}$↓      &0.077 & 0.208 & 0.155 & 0.190 & 0.167 & 0.138 & 0.133 & 0.119 & 0.132\\ 
\hline
\multirow{8}*{\rotatebox{90}{\emph{STERE} 
\cite{49niu2012leveraging}}} &$S_\alpha$↑&0.692 & 0.660 & 0.731 & 0.728 & 0.708 & 0.757 & 0.825 & 0.848 & 0.873 \\
 &max $F_\beta$↑   &0.669 & 0.633 & 0.740 & 0.719 & 0.755 & 0.757 & 0.823 & 0.831 & 0.863 \\
 &mean $F_\beta$↑  &0.478 & 0.501 & 0.590 & 0.527 & 0.610 & 0.617 & 0.806 & 0.758 & 0.813 \\
 &adp $F_\beta$↑   &0.661 & 0.595 & 0.742 & 0.744 & 0.748 & 0.742 & 0.807 & 0.771 & 0.829 \\
 &max $E_\xi$↑   &0.806 & 0.787 & 0.819 & 0.809 & 0.846 & 0.847 & 0.887 & 0.912 & 0.927 \\
 &mean $E_\xi$↑  &0.592 & 0.601 & 0.655 & 0.614 & 0.665 & 0.691 & 0.872 & 0.841 & 0.873 \\
 &adp $E_\xi$↑   &0.793 & 0.749 & 0.831 & 0.830 & 0.825 & 0.838 & 0.886 & 0.864 & 0.901 \\
 &$\mathcal{M}$↓      &0.200 & 0.250 & 0.148 & 0.176 & 0.143 & 0.141 & 0.075 & 0.086 & 0.068\\ 
\hline
\end{tabular}
}
\label{tab:addlabel}
\end{table}

\begin{table}[H]
\centering
\resizebox{\textwidth}{!}
{
\begin{tabular}{ccccccc|cccc}
\hline
\hline
 & &ACSD&LBE&DCMC&MDSF&SE&DF&AFNet&CTMF&MMCI\\
 &Metric&ICIP14&CVPR16&SPL16&TIP17&ICME16&TIP17&arXiv19&TOC18&PR19\\
 & &[52]&[19]&[53]&[54]&[55]&[64]&[56]&[57]&[9]\\
\hline
\hline
\multirow{8}*{\rotatebox{90}{\emph{DES} 
\cite{18cheng2014depth}}} &$S_\alpha$↑      &0.728 & 0.703 & 0.707 & 0.741 & 0.741 & 0.752 & 0.770 & 0.863 & 0.848 \\ 
 &max $F_\beta$↑   &0.756 & 0.788 & 0.666 & 0.746 & 0.741 & 0.766 & 0.728 & 0.844 & 0.822 \\
 &mean $F_\beta$↑  &0.513 & 0.576 &  0.542 & 0.523 & 0.617 & 0.604 & 0.713 & 0.756 & 0.735 \\
 &adp $F_\beta$↑   &0.717 & 0.796 & 0.702 & 0.744 & 0.726 & 0.753 & 0.730 & 0.778 & 0.762 \\
 &max $E_\xi$↑   &0.850 & 0.890 & 0.773 & 0.851 & 0.856 & 0.870 & 0.881 & 0.932 & 0.928 \\
 &mean $E_\xi$↑  &0.612 & 0.649 & 0.632 & 0.621 & 0.707 & 0.684 & 0.810 & 0.826 & 0.825 \\
 &adp $E_\xi$↑   &0.855 & 0.911 & 0.849 & 0.869 & 0.852 & 0.877 & 0.874 & 0.911 & 0.904 \\
 &$\mathcal{M}$↓      &0.169 & 0.208 & 0.111 & 0.122 & 0.090 & 0.093 & 0.068 & 0.055 & 0.065\\ 
\hline
\multirow{8}*{\rotatebox{90}{\emph{NLPR} 
\cite{39zhai2021bifurcated}}} &$S_\alpha$↑      &0.673 & 0.762 & 0.724 & 0.805 & 0.756 & 0.802 & 0.799 & 0.860 & 0.856 \\
 &max $F_\beta$↑   &0.673 & 0.762 & 0.724 & 0.805 & 0.756 & 0.802 & 0.799 & 0.860 & 0.856 \\
 &mean $F_\beta$↑  &0.429 & 0.626 & 0.543 & 0.649 & 0.624 & 0.664 & 0.755 & 0.740 & 0.737 \\
 &adp $F_\beta$↑   &0.535 & 0.736 & 0.614 & 0.665 & 0.692 & 0.744 & 0.747 & 0.724 & 0.730 \\
 &max $E_\xi$↑   &0.780 & 0.855 & 0.793 & 0.885 & 0.847 & 0.880 & 0.879 & 0.929 & 0.913 \\
 &mean $E_\xi$↑  &0.578 & 0.719 & 0.684 & 0.745 & 0.742 & 0.775 & 0.851 & 0.840 & 0.841 \\
 &adp $E_\xi$↑   &0.742 & 0.855 & 0.786 & 0.812 & 0.839 & 0.868 & 0.884 & 0.869 & 0.872 \\
 &$\mathcal{M}$↓      &0.179 & 0.081 & 0.117 & 0.095 & 0.091 & 0.085 & 0.058 & 0.056 & 0.059\\ 
\hline
\multirow{8}*{\rotatebox{90}{\emph{SSD} 
\cite{51zhu2017three}}} &$S_\alpha$↑      &0.675 & 0.621 & 0.704 & 0.673 & 0.675 & 0.747 & 0.714 & 0.776 & 0.813 \\
 &max $F_\beta$↑   &0.682 & 0.619 & 0.711 & 0.703 & 0.710 & 0.735 & 0.687 & 0.729 & 0.781 \\
 &mean $F_\beta$↑  &0.469 & 0.489 & 0.572 & 0.470 & 0.564 & 0.624 & 0.672 & 0.689 & 0.721 \\
 &adp $F_\beta$↑   &0.656 & 0.613 & 0.679 & 0.674 & 0.693 & 0.724 & 0.694 & 0.710 & 0.748 \\
 &max $E_\xi$↑   &0.785 & 0.736 & 0.786 & 0.779 & 0.800 & 0.828 & 0.807 & 0.865 & 0.882 \\
 &mean $E_\xi$↑  &0.566 & 0.574 & 0.646 & 0.576 & 0.631 & 0.690 & 0.762 & 0.796 & 0.796 \\
 &adp $E_\xi$↑   &0.765 & 0.729 & 0.786 & 0.772 & 0.778 & 0.812 & 0.803 & 0.838 & 0.860 \\
 &$\mathcal{M}$↓      &0.203 & 0.278 & 0.169 & 0.192 & 0.165 & 0.142 & 0.118 & 0.099 & 0.082\\ 
\hline
\multirow{8}*{\rotatebox{90}{\emph{SIP} 
\cite{50fan2020rethinking}}} &$S_\alpha$↑      &0.732 & 0.727 & 0.683 & 0.717 & 0.628 & 0.653 & 0.720 & 0.716 & 0.833 \\
 &max $F_\beta$↑   &0.763 & 0.751 & 0.618 & 0.698 & 0.661 & 0.657 & 0.712 & 0.694 & 0.818 \\
 &mean $F_\beta$↑  &0.542 & 0.571 & 0.499 & 0.568 & 0.515 & 0.464 & 0.702 & 0.608 & 0.771 \\
 &adp $F_\beta$↑   &0.727 & 0.733 & 0.645 & 0.694 & 0.662 & 0.673 & 0.705 & 0.684 & 0.795 \\
 &max $E_\xi$↑   &0.838 & 0.853 & 0.743 & 0.798 & 0.771 & 0.759 & 0.819 & 0.829 & 0.897 \\
 &mean $E_\xi$↑  &0.614 & 0.651 & 0.598 & 0.645 & 0.592 & 0.565 & 0.793 & 0.705 & 0.845 \\
 &adp $E_\xi$↑   &0.827 & 0.841 & 0.786 & 0.805 & 0.756 & 0.794 & 0.815 & 0.824 & 0.886 \\
 &$\mathcal{M}$↓      &0.172 & 0.200 & 0.186 & 0.167 & 0.164 & 0.185 & 0.118 & 0.139 & 0.086\\ 
\hline
\end{tabular}
}
\end{table}

\begin{table}[H]
\centering
\caption{Continuation of Tab. 2}
\resizebox{\textwidth}{!}
{
\begin{tabular}{ccccccccc}
\hline
\hline
 & &PCF&TANet&CPFP&DMRA&D3Net&MCI-Net&Ours\\
 &Metric&CVPR18&TIP19	CVPR19&ICCV19&TNNLS20&Neurocomputing21&21\\
 & &[7]&[38]&[6]&[5]&[50]&[58]& \\
\hline
\hline
\multirow{8}*{\rotatebox{90}{\emph{NJU2K} 
\cite{48fan2014salient}}} &$S_\alpha$↑ &0.877 & 0.878 & 0.879 & 0.886 & 0.893 & 0.900 & \textbf{0.912} \\ 
 &max $F_\beta$↑   &0.872 & 0.874 & 0.877 & 0.886 & 0.887 & -     & \textbf{0.913} \\
 &mean $F_\beta$↑  &0.840 & 0.841 & 0.850 & 0.873 & 0.859 & 0.873 & \textbf{0.904} \\
 &adp $F_\beta$↑   &0.844 & 0.844 & 0.837 & 0.872 & 0.840 & -     & \textbf{0.904} \\
 &max $E_\xi$↑     &0.924 & 0.925 & 0.926 & 0.927 & 0.930 & -     & \textbf{0.942} \\
 &mean $E_\xi$↑    &0.895 & 0.895 & 0.910 & 0.920 & 0.910 & 0.920 & \textbf{0.936} \\
 &adp $E_\xi$↑     &0.896 & 0.893 & 0.895 & 0.908 & 0.894 & -     & \textbf{0.917} \\
 &$\mathcal{M}$↓   &0.059 & 0.060 & 0.053 & 0.051 & 0.051 & 0.050 & \textbf{0.038} \\ 
\hline
\multirow{8}*{\rotatebox{90}{\emph{LFSD} 
\cite{15li2014saliency}}} &$S_\alpha$↑ &0.794 & 0.801 & 0.828 & 0.847          & 0.825 & - & \textbf{0.862} \\
 &max $F_\beta$↑   &0.779 & 0.796 & 0.826 & 0.856          & 0.810 & - & \textbf{0.863} \\
 &mean $F_\beta$↑  &-     & 0.771 & 0.811 & \textbf{0.845} & 0.796 & - & 0.838          \\
 &adp $F_\beta$↑   &-     & 0.794 & 0.813 & 0.849          & 0.805 & - & \textbf{0.851} \\
 &max $E_\xi$↑     &-     & 0.847 & 0.872 & \textbf{0.900} & 0.862 & - & 0.898          \\
 &mean $E_\xi$↑    &0.835 & 0.821 & 0.863 & \textbf{0.893} & 0.849 & - & 0.873          \\
 &adp $E_\xi$↑     &-     & 0.845 & 0.867 & \textbf{0.899} & 0.853 & - & 0.885          \\
 &$\mathcal{M}$↓   &0.112 & 0.111 & 0.088 & \textbf{0.075} & 0.095 & - & 0.077 \\ 

\hline
\multirow{8}*{\rotatebox{90}{\emph{STERE} 
\cite{49niu2012leveraging}}} &$S_\alpha$↑ &0.875 & 0.871 & 0.879 & 0.835 & 0.889          & 0.901          & \textbf{0.904} \\
 &max $F_\beta$↑   &0.860 & 0.861 & 0.874 & 0.847 & \textbf{0.878} & -              & 0.780          \\
 &mean $F_\beta$↑  &0.818 & 0.828 & 0.841 & 0.837 & 0.841          & 0.872          & \textbf{0.892} \\
 &adp $F_\beta$↑   &0.826 & 0.835 & 0.830 & 0.844 & 0.829          & -              & \textbf{0.891} \\
 &max $E_\xi$↑     &0.925 & 0.923 & 0.925 & 0.911 & 0.929          & -              & \textbf{0.936} \\
 &mean $E_\xi$↑    &0.887 & 0.893 & 0.912 & 0.879 & 0.906          & 0.929          & \textbf{0.929} \\
 &adp $E_\xi$↑     &0.897 & 0.906 & 0.903 & 0.900 & 0.902          & -              & \textbf{0.921} \\
 &$\mathcal{M}$↓   &0.064 & 0.060 & 0.051 & 0.066 & 0.054          & \textbf{0.042} & 0.045  \\ 

\hline
\multirow{8}*{\rotatebox{90}{\emph{DES} 
\cite{18cheng2014depth}}} &$S_\alpha$↑ &0.842 & 0.858 & 0.872 & 0.900 & 0.898 & 0.927          & \textbf{0.931} \\
 &max $F_\beta$↑   &0.804 & 0.827 & 0.846 & 0.888 & 0.880 & -              & \textbf{0.923} \\
 &mean $F_\beta$↑  &0.765 & 0.790 & 0.824 & 0.873 & 0.851 & 0.897          & \textbf{0.903} \\
 &adp $F_\beta$↑   &0.782 & 0.795 & 0.829 & 0.866 & 0.863 & -              & \textbf{0.903} \\
 &max $E_\xi$↑     &0.893 & 0.910 & 0.923 & 0.943 & 0.935 & -              & \textbf{0.967} \\
 &mean $E_\xi$↑    &0.838 & 0.863 & 0.889 & 0.933 & 0.902 & \textbf{0.957} & 0.944          \\
 &adp $E_\xi$↑     &0.912 & 0.919 & 0.927 & 0.944 & 0.946 & -              & \textbf{0.967} \\
 &$\mathcal{M}$↓   &0.049 & 0.046 & 0.038 & 0.030 & 0.033 & 0.024          & \textbf{0.023}  \\ 

\hline
\multirow{8}*{\rotatebox{90}{\emph{NLPR} 
\cite{39zhai2021bifurcated}}} &$S_\alpha$↑ &0.874 & 0.886 & 0.888 & 0.899 & 0.905 & 0.917 & \textbf{0.931} \\
 &max $F_\beta$↑   &0.841 & 0.863 & 0.867 & 0.879 & 0.885 & -     & \textbf{0.918} \\
 &mean $F_\beta$↑  &0.802 & 0.819 & 0.840 & 0.864 & 0.852 & 0.890 & \textbf{0.894} \\
 &adp $F_\beta$↑   &0.795 & 0.796 & 0.823 & 0.854 & 0.832 & -     & \textbf{0.880} \\
 &max $E_\xi$↑     &0.925 & 0.941 & 0.932 & 0.947 & 0.945 & -     & \textbf{0.962} \\
 &mean $E_\xi$↑    &0.887 & 0.902 & 0.918 & 0.940 & 0.923 & 0.947 & \textbf{0.947} \\
 &adp $E_\xi$↑     &0.916 & 0.916 & 0.924 & 0.941 & 0.931 & -     & \textbf{0.954} \\
 &$\mathcal{M}$↓   &0.044 & 0.041 & 0.036 & 0.031 & 0.033 & 0.027 & \textbf{0.025}\\ 

\hline
\end{tabular}
}
\label{tab:addlabel}
\end{table}

\begin{table}[H]
\centering
\resizebox{\textwidth}{!}
{
\begin{tabular}{ccccccccc}
\hline
\hline
 & &PCF&TANet&CPFP&DMRA&D3Net&MCI-Net&Ours\\
 &Metric&CVPR18&TIP19	CVPR19&ICCV19&TNNLS20&Neurocomputing21&21\\
 & &[7]&[38]&[6]&[5]&[50]&[58]& \\
\hline
\hline
\multirow{8}*{\rotatebox{90}{\emph{SSD} 
\cite{51zhu2017three}}} &$S_\alpha$↑ &0.841 & 0.839 & 0.807 & 0.857 & 0.865 & 0.860          & \textbf{0.880} \\
 &max $F_\beta$↑   &0.807 & 0.810 & 0.766 & 0.844 & 0.846 & -              & \textbf{0.868} \\
 &mean $F_\beta$↑  &0.777 & 0.773 & 0.747 & 0.828 & 0.815 & 0.820          & \textbf{0.843} \\
 &adp $F_\beta$↑   &0.791 & 0.767 & 0.726 & 0.821 & 0.790 & -              & \textbf{0.836} \\
 &max $E_\xi$↑     &0.894 & 0.897 & 0.852 & 0.906 & 0.907 & -              & \textbf{0.918} \\
 &mean $E_\xi$↑    &0.856 & 0.861 & 0.839 & 0.897 & 0.886 & 0.901          & \textbf{0.902} \\
 &adp $E_\xi$↑     &0.886 & 0.879 & 0.832 & 0.892 & 0.885 & -              & \textbf{0.901} \\
 &$\mathcal{M}$↓   &0.062 & 0.063 & 0.082 & 0.058 & 0.059 & 0.052          & \textbf{0.050} \\

\hline
\multirow{8}*{\rotatebox{90}{\emph{SIP} 
\cite{50fan2020rethinking}}} &$S_\alpha$↑ &0.842 & 0.835 & 0.850 & 0.806 & 0.864 & 0.867          & \textbf{0.879} \\
 &max $F_\beta$↑   &0.838 & 0.830 & 0.851 & 0.821 & 0.861 & -              & \textbf{0.884} \\
 &mean $F_\beta$↑  &0.814 & 0.803 & 0.821 & 0.811 & 0.830 & 0.840          & \textbf{0.867} \\
 &adp $F_\beta$↑   &0.825 & 0.809 & 0.819 & 0.819 & 0.829 & -              & \textbf{0.870} \\
 &max $E_\xi$↑     &0.901 & 0.895 & 0.903 & 0.875 & 0.910 & -              & \textbf{0.920} \\
 &mean $E_\xi$↑    &0.878 & 0.870 & 0.893 & 0.844 & 0.893 & \textbf{0.909} & 0.903          \\
 &adp $E_\xi$↑     &0.899 & 0.893 & 0.899 & 0.863 & 0.901 & -              & \textbf{0.914} \\
 &$\mathcal{M}$↓   &0.071 & 0.075 & 0.064 & 0.085 & 0.063 & 0.056          & \textbf{0.056}  \\ 

\hline
\end{tabular}
}
\label{tab:addlabel}
\end{table}

Moreover, we present the F-measure of our method and other competitive methods in Fig. 6. Our method not only produces stable F-measure but also achieves the best performance in all 14 methods. To further evaluate our performance, we follow D3Net 
[50] and compute the PR curves comparing different methods on 7 datasets. The experimental results are presented in Fig. 7, which further give evidence of effectiveness of our proposed method.

\begin{figure}[H]
	\centering
	\subfigure[F-measure curves on \emph{NJU2K}\cite{48fan2014salient}]{
		\includegraphics[width=0.75\textwidth]{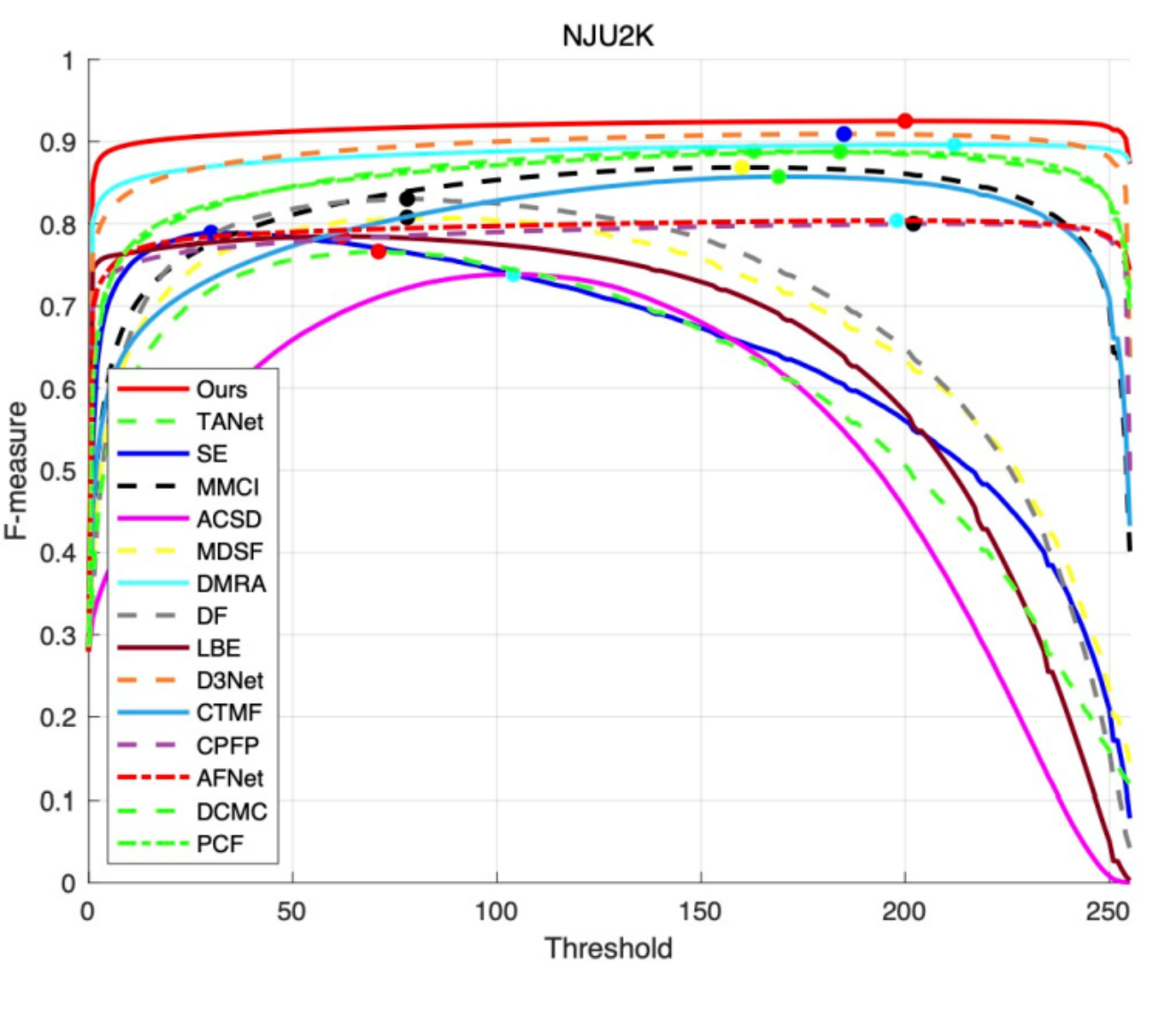}
	}
     \\
     \subfigure[F-measure curves on \emph{NLPR}\cite{39zhai2021bifurcated}]{
		\includegraphics[width=0.75\textwidth]{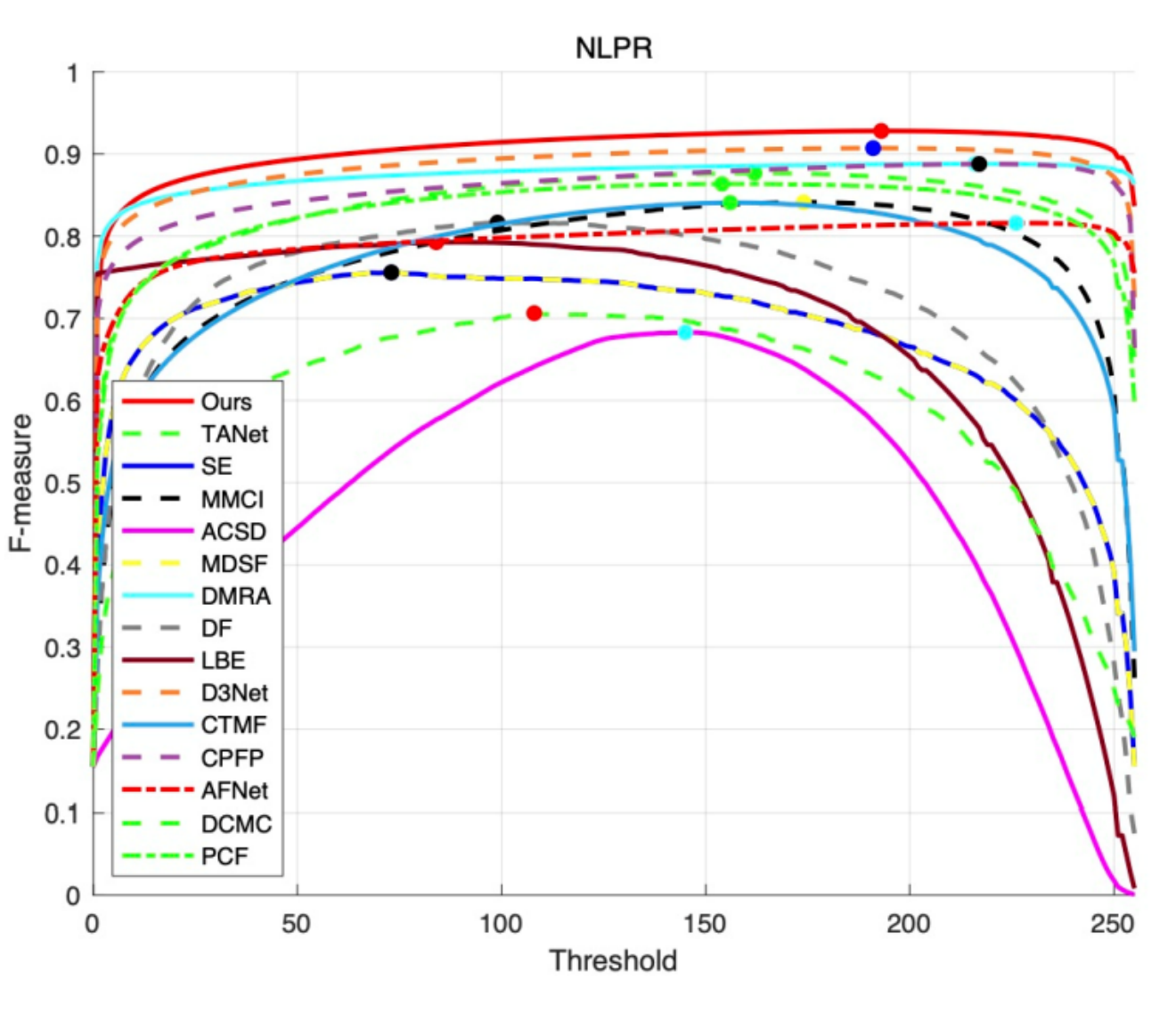}
	}
\end{figure}
\addtocounter{figure}{-1}
\begin{figure}[H]
\addtocounter{figure}{1}
	\centering
	\subfigure[F-measure curves on \emph{LFSD}\cite{15li2014saliency}]{
		\includegraphics[width=0.75\textwidth]{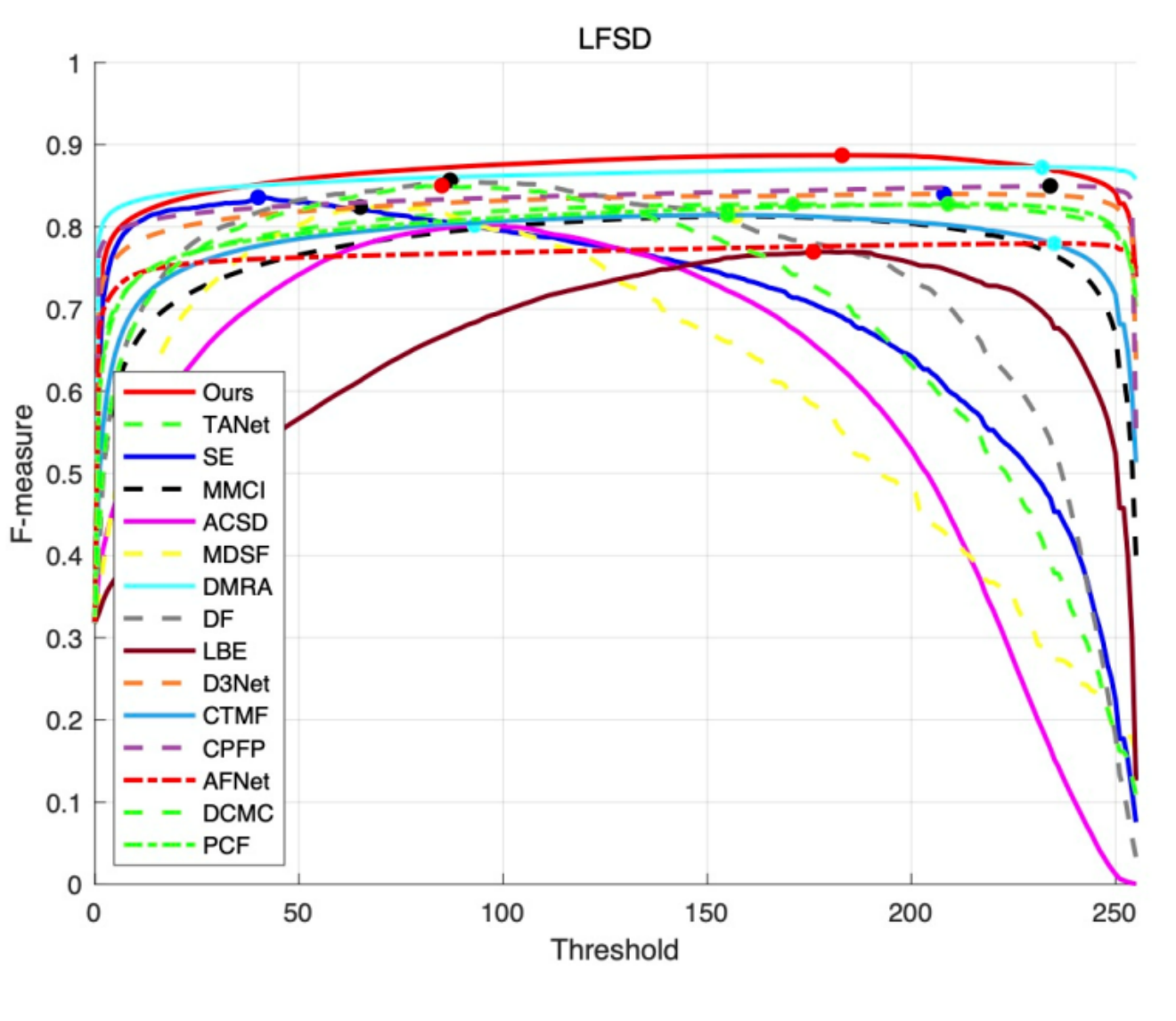}
	}
	\subfigure[F-measure curves on \emph{DES}\cite{18cheng2014depth}]{
		\includegraphics[width=0.75\textwidth]{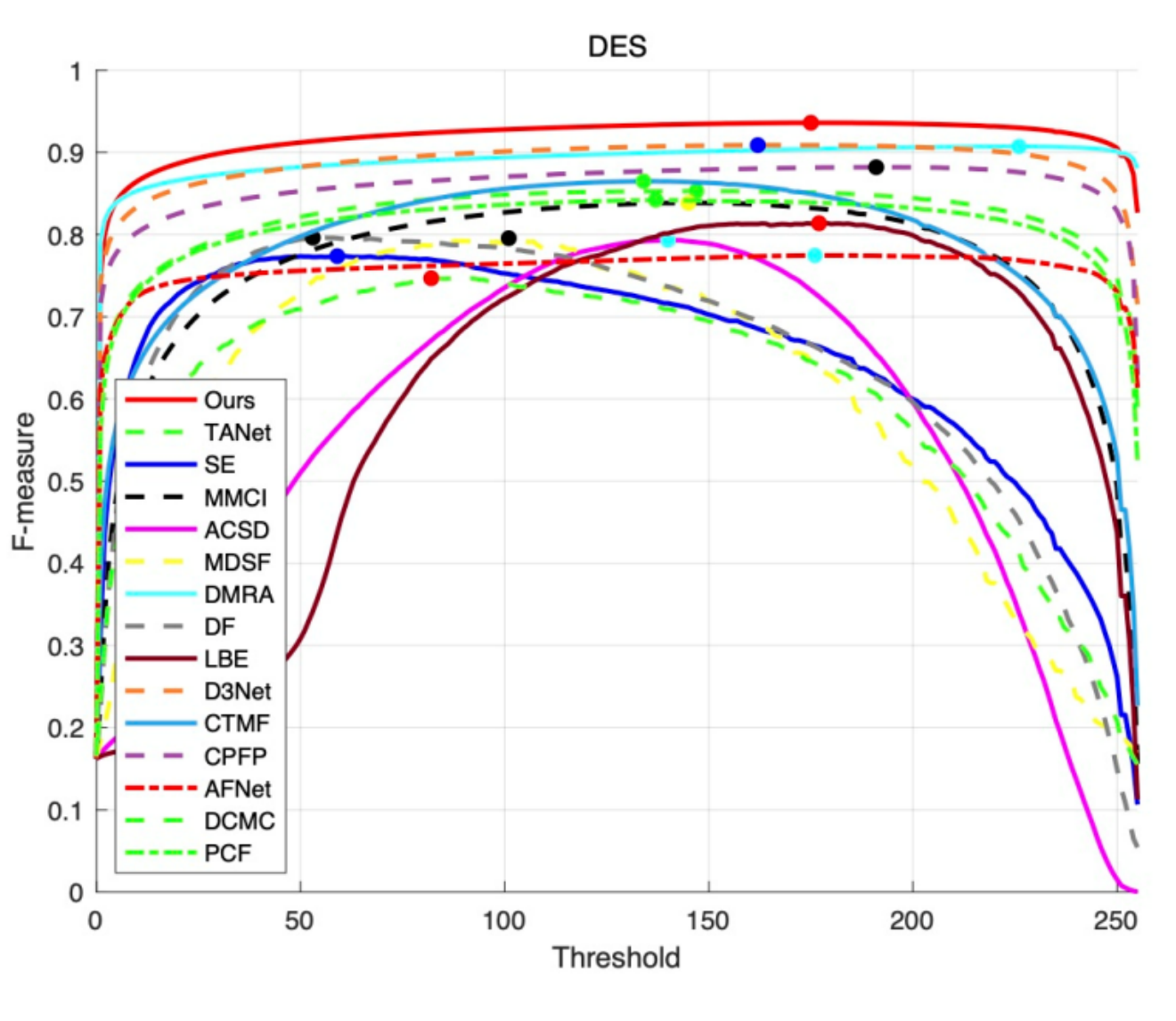}
	}
\end{figure}
\addtocounter{figure}{-1}
\begin{figure}[H]
\addtocounter{figure}{1}
	\centering
	\subfigure[F-measure curves on \emph{SSD}\cite{51zhu2017three}]{
		\includegraphics[width=0.75\textwidth]{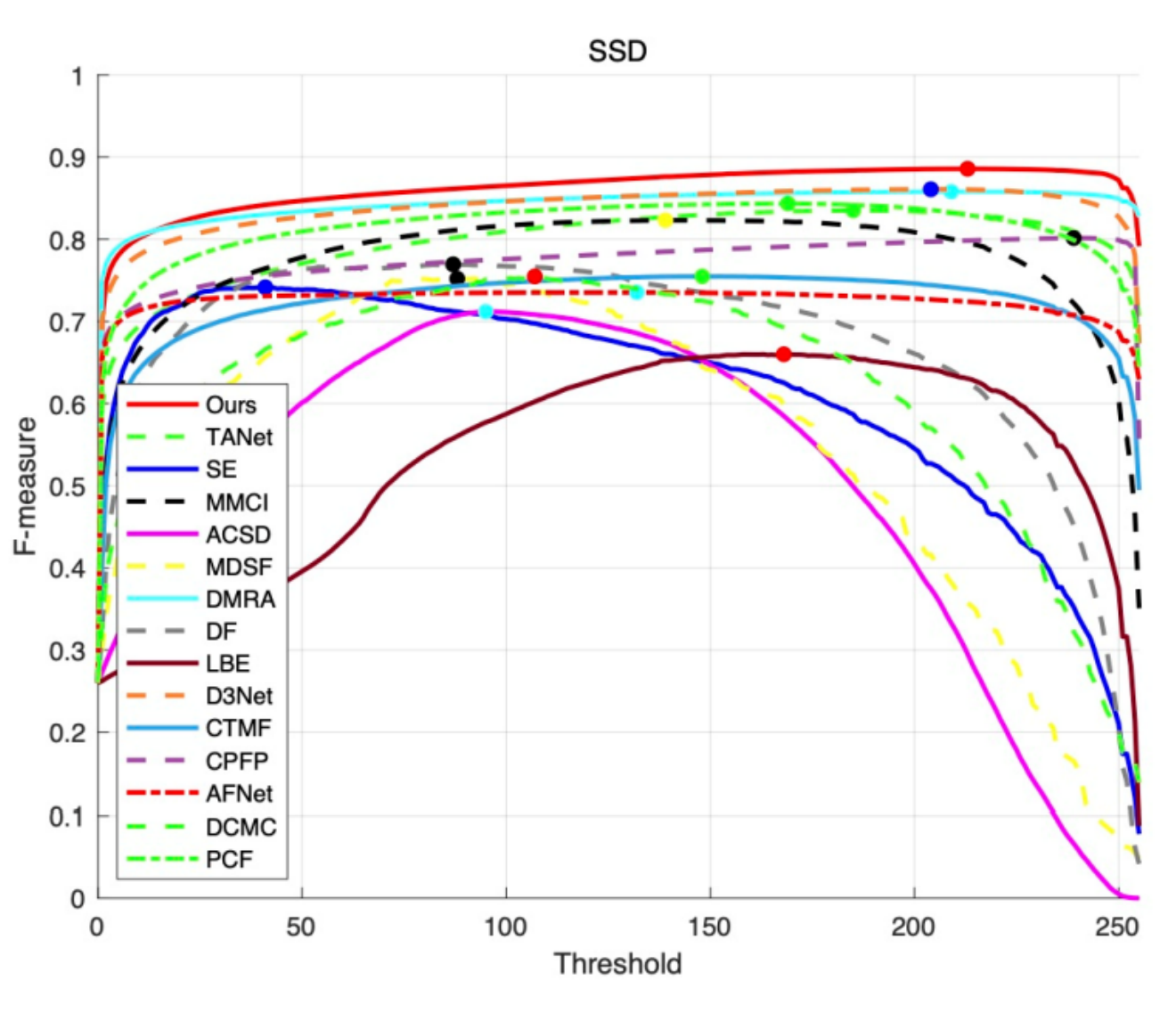}
	}
	\\
	\subfigure[F-measure curves on \emph{SIP}\cite{50fan2020rethinking}]{
		\includegraphics[width=0.75\textwidth]{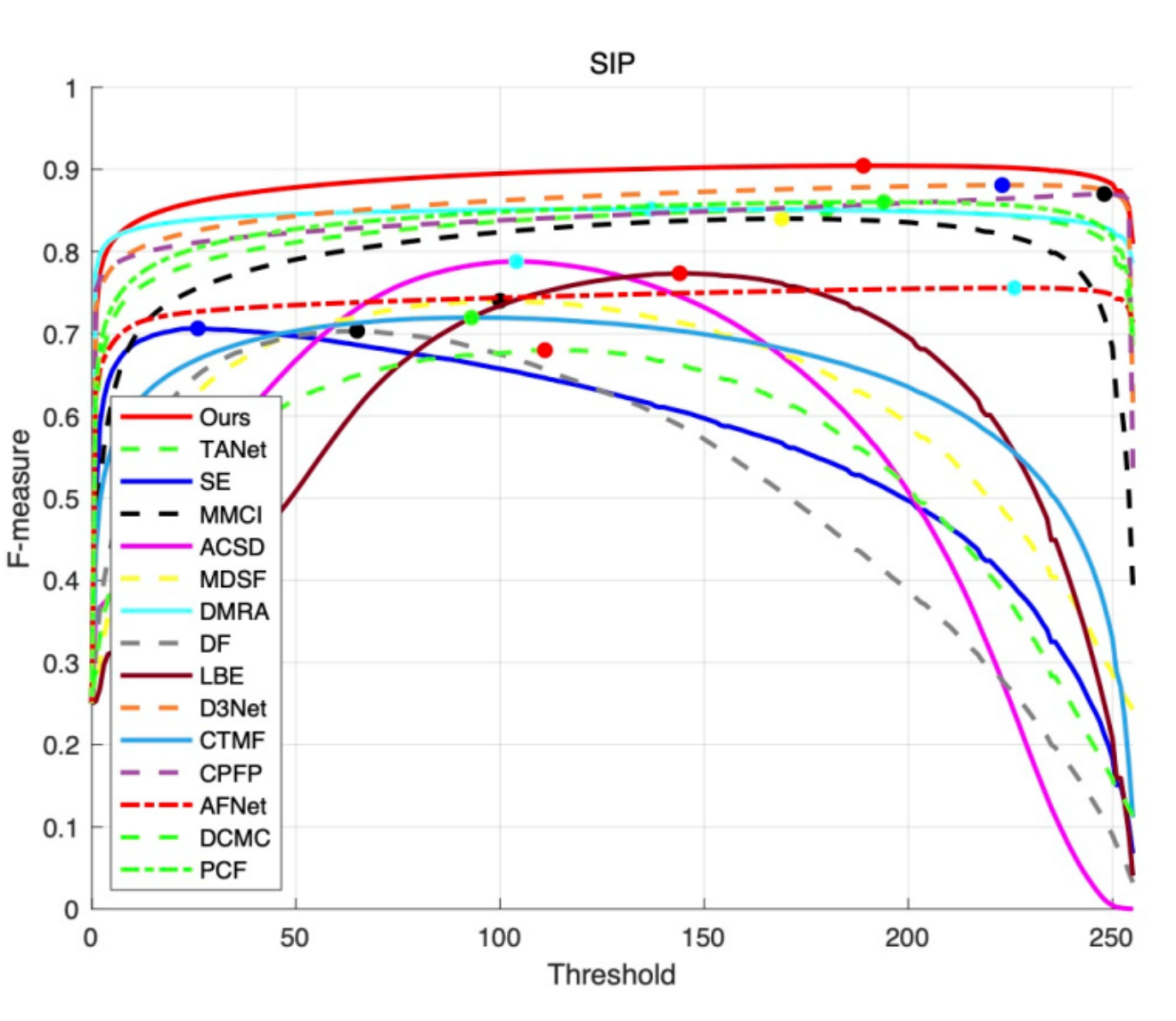}
	}
\end{figure}
\addtocounter{figure}{-1}
\begin{figure}[H]
\addtocounter{figure}{1}
	\centering
	\subfigure[F-measure curves on \emph{STERE}\cite{49niu2012leveraging}]{
	\includegraphics[width=0.75\textwidth]{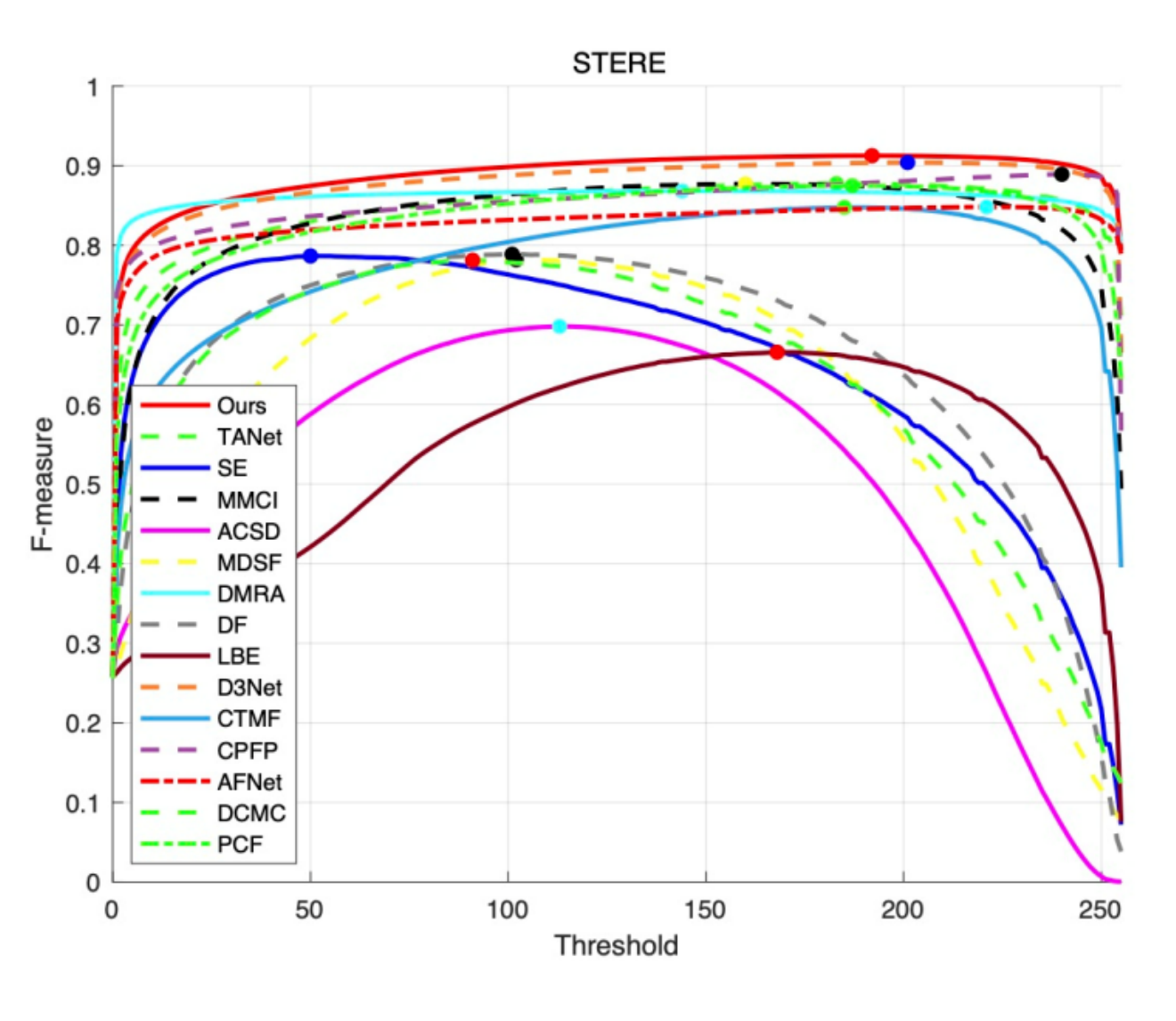}

	}
	\caption{Comparison of F-measure curves between our method and 14 state-of-the-art methods on 7 datasets. Compared to other state-of-art methods, our MSIRN method achieves favorable performance on the majority of the data sets.}
     \label{fig6}
\end{figure}

\begin{figure}[H]
	\centering
	\subfigure[PR curves on \emph{NJU2K}\cite{48fan2014salient}]{
		\includegraphics[width=0.75\textwidth]{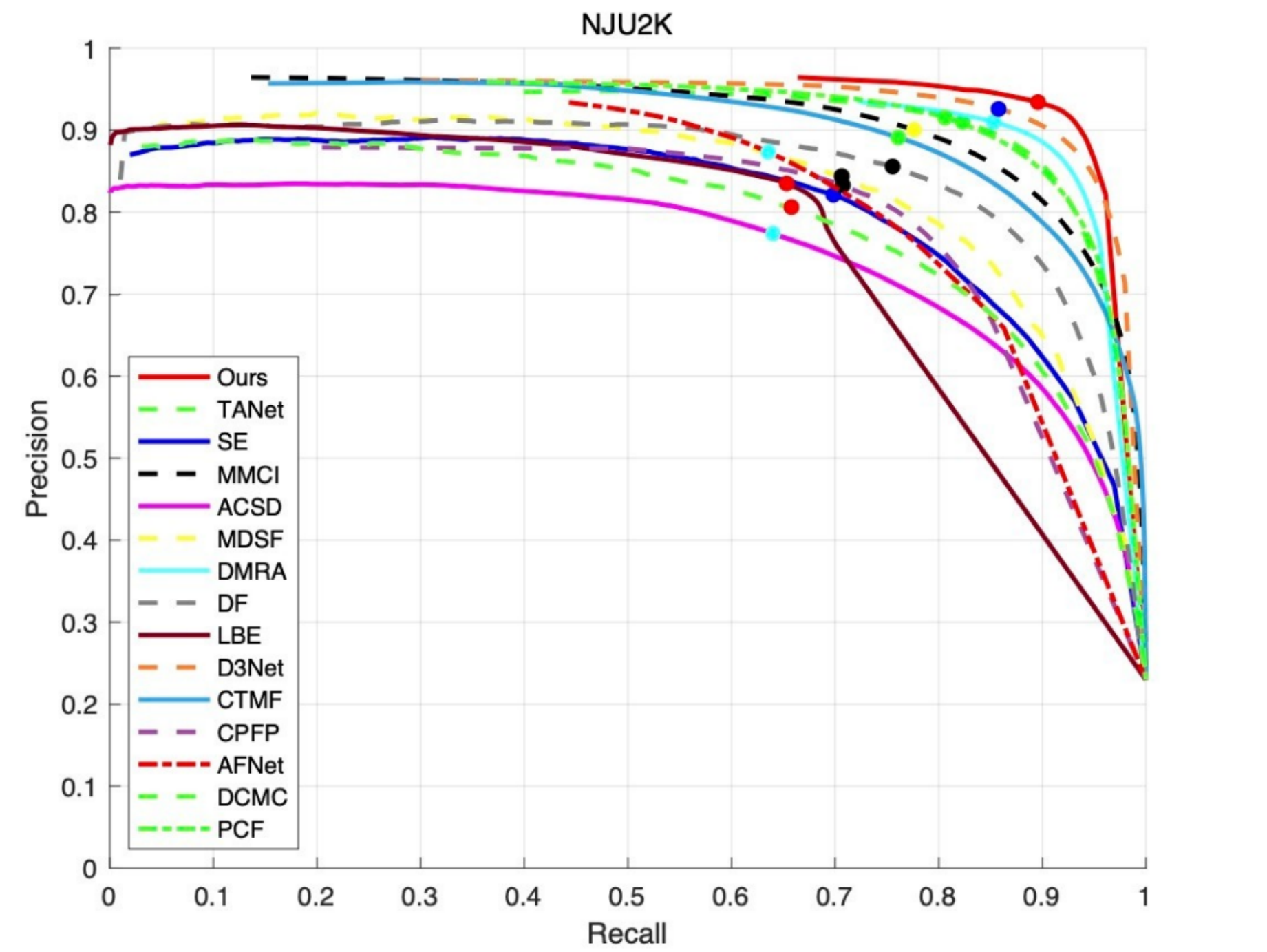}
	}
\end{figure}
\addtocounter{figure}{-1}
\begin{figure}[H]
\addtocounter{figure}{1}
	\centering
     \subfigure[PR curves on \emph{NLPR}\cite{39zhai2021bifurcated}]{
		\includegraphics[width=0.75\textwidth]{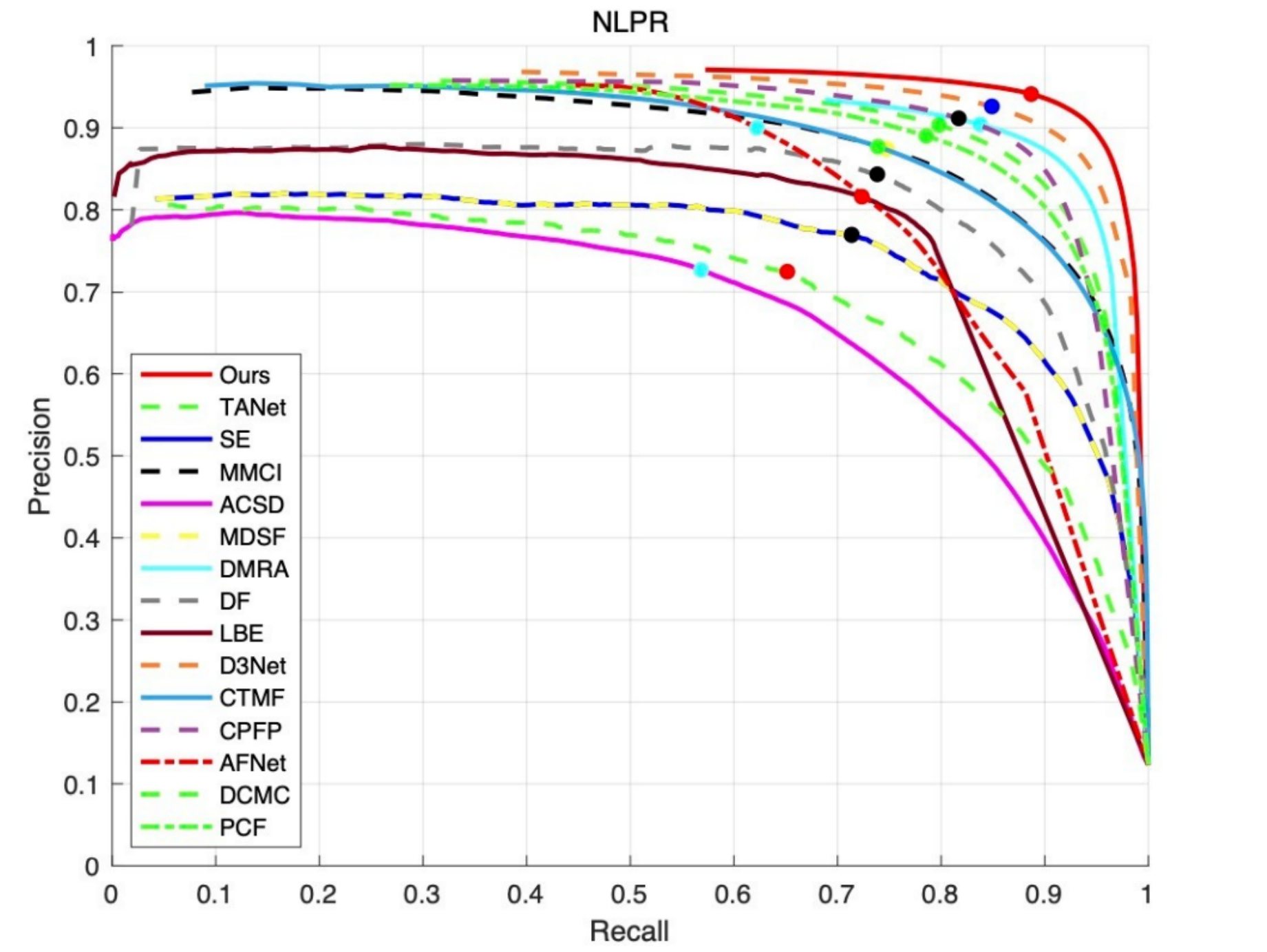}
	}
	\subfigure[PR curves on \emph{LFSD}\cite{15li2014saliency}]{
		\includegraphics[width=0.75\textwidth]{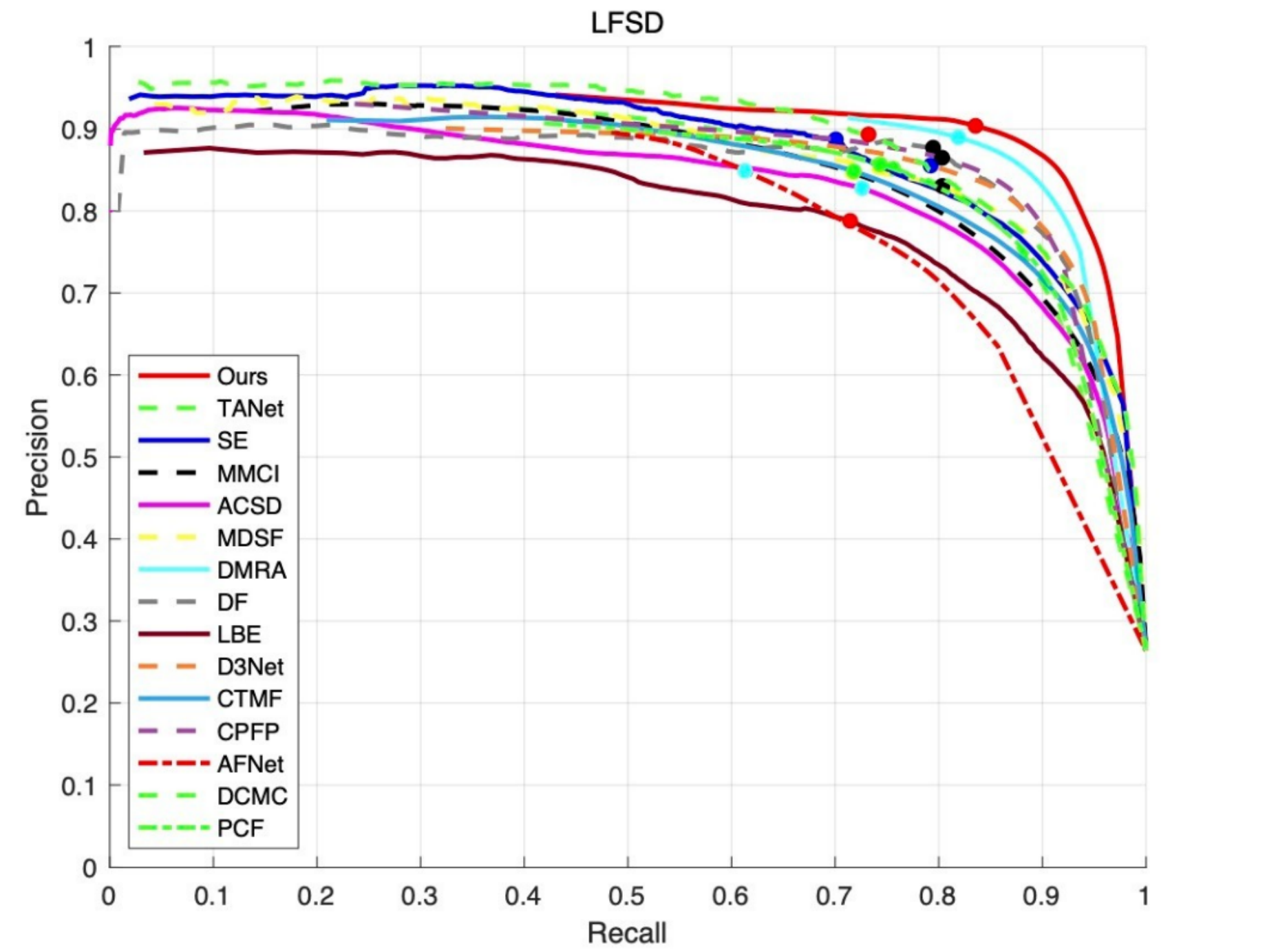}
	}
\end{figure}
\addtocounter{figure}{-1}
\begin{figure}[H]
\addtocounter{figure}{1}
	\centering
	\subfigure[PR curves on \emph{DES}\cite{18cheng2014depth}]{
		\includegraphics[width=0.75\textwidth]{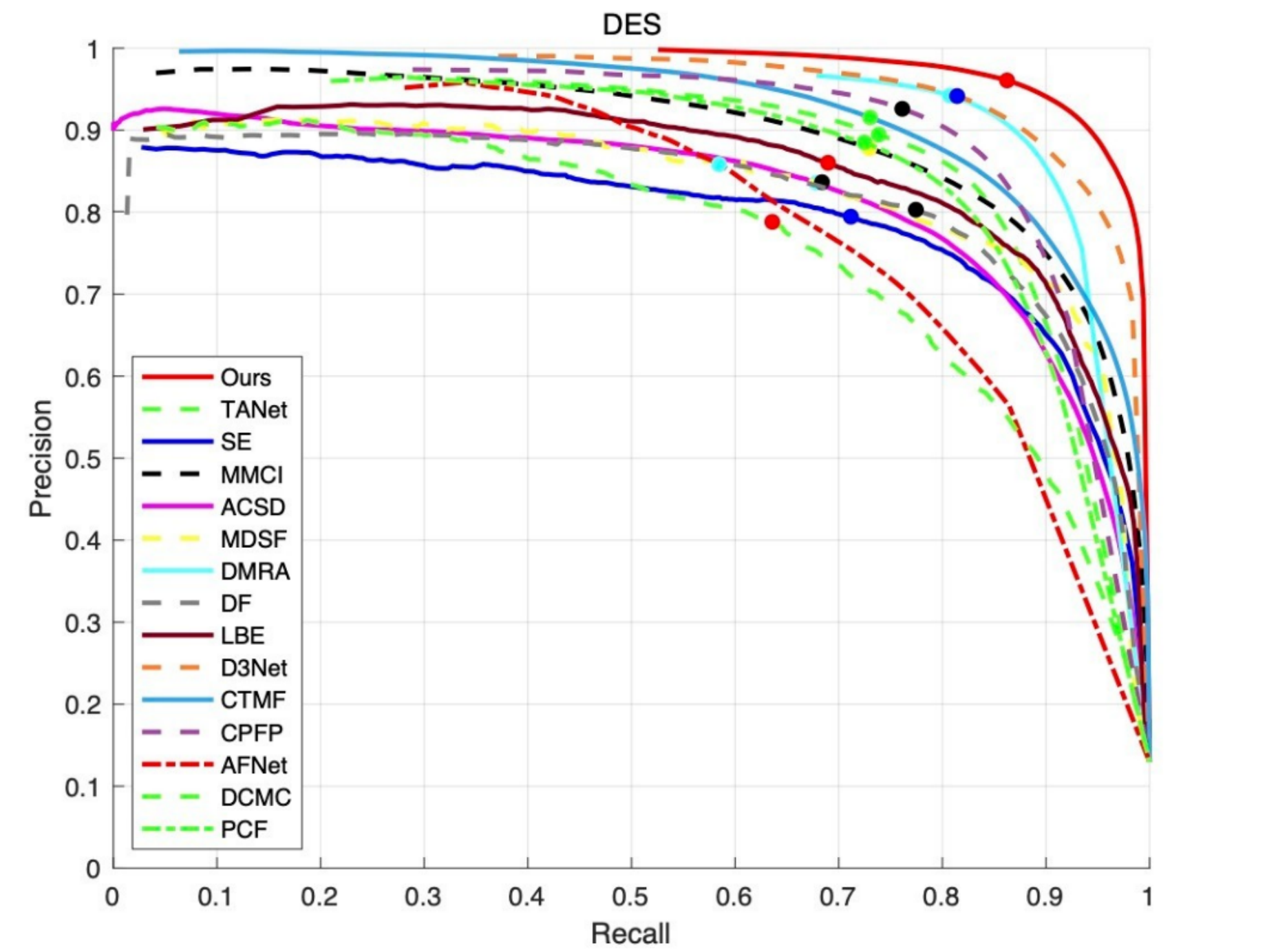}
	}

	\subfigure[PR curves on \emph{SSD}\cite{51zhu2017three}]{
		\includegraphics[width=0.75\textwidth]{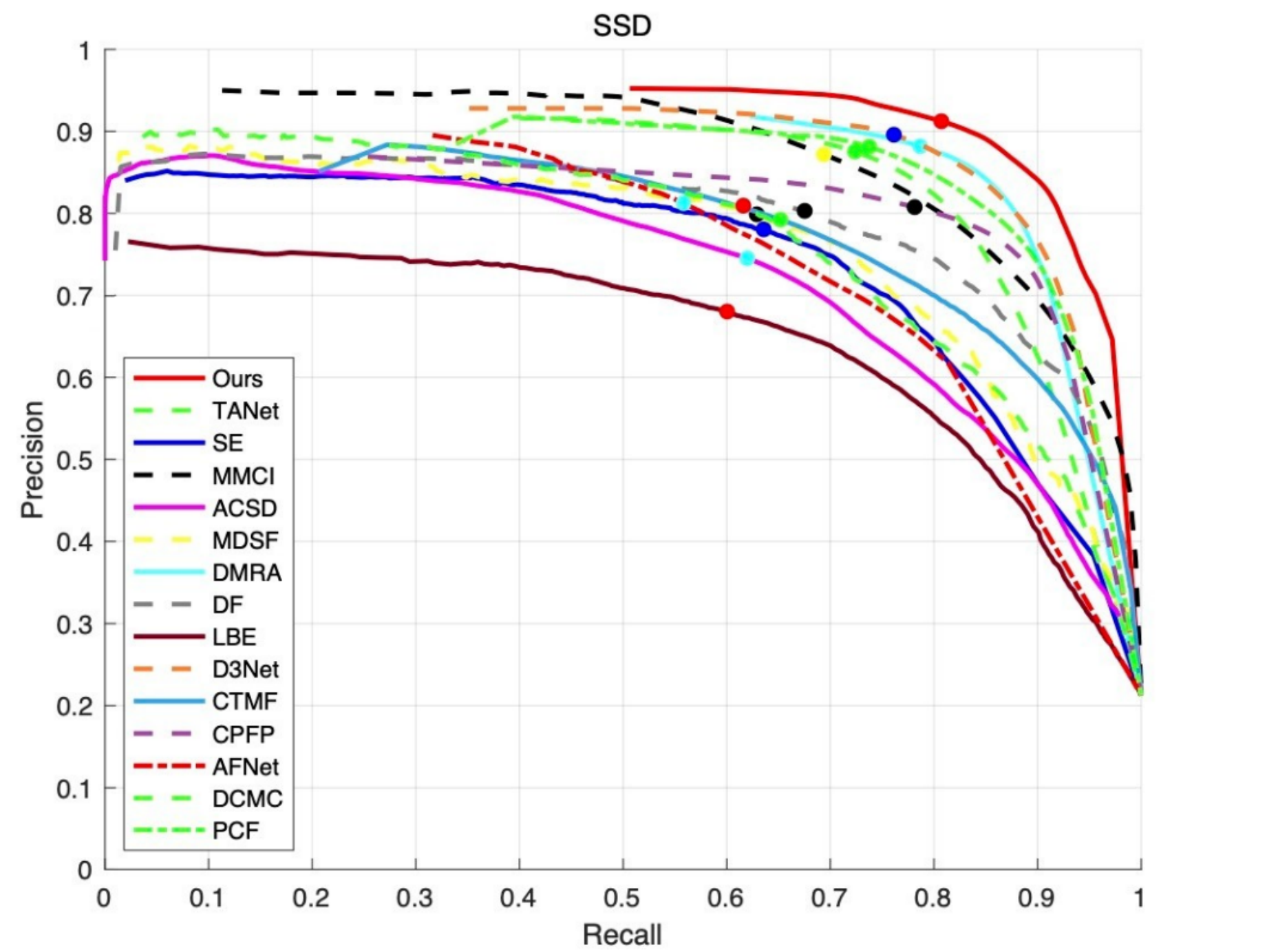}
	}
\end{figure}
\addtocounter{figure}{-1}
\begin{figure}[H]
\addtocounter{figure}{1}
	\centering
	\subfigure[PR curves on \emph{SIP}\cite{50fan2020rethinking}]{
		\includegraphics[width=0.75\textwidth]{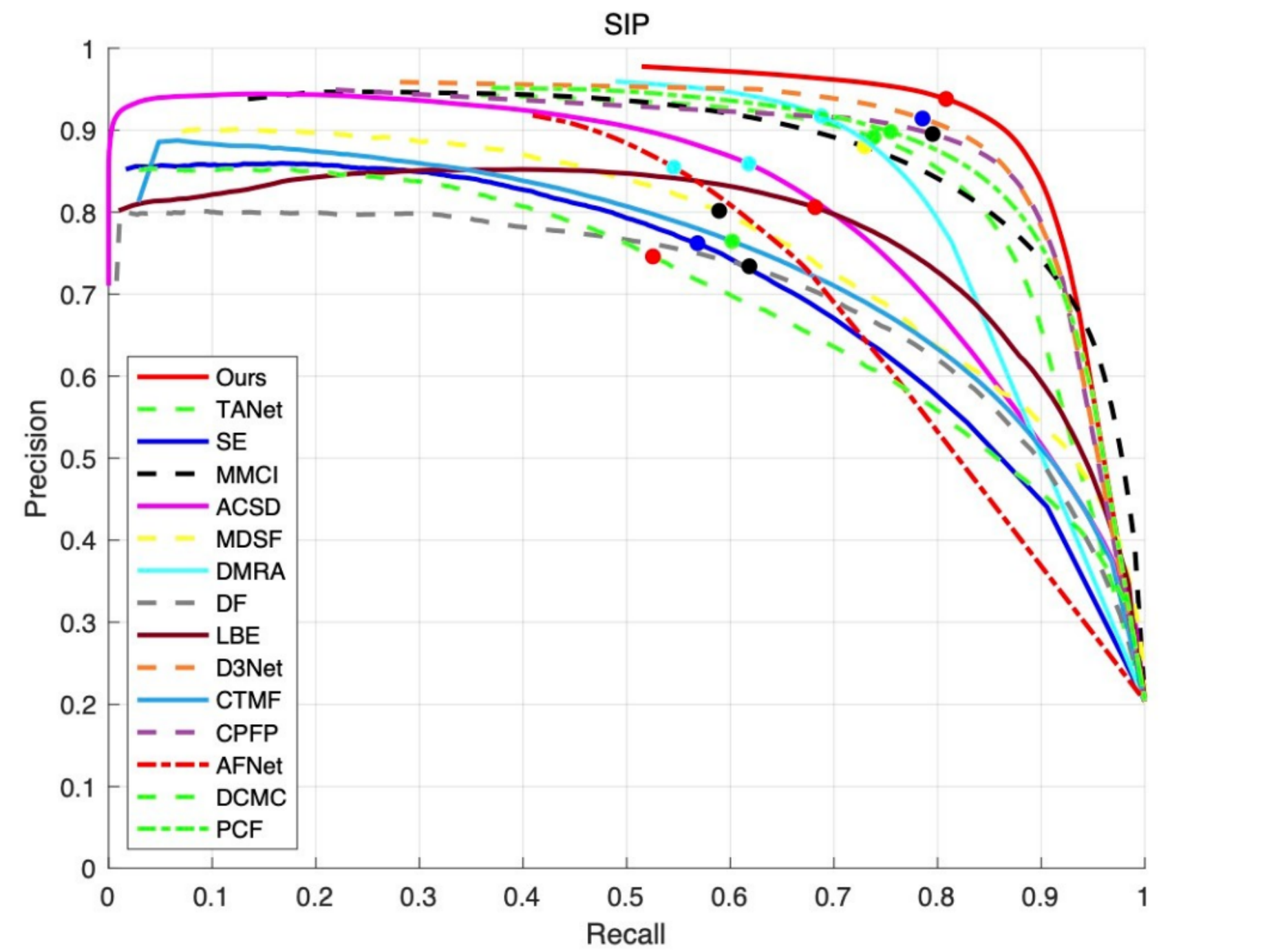}
	}

	\subfigure[PR curves on \emph{STERE}\cite{49niu2012leveraging}]{
	\includegraphics[width=0.75\textwidth]{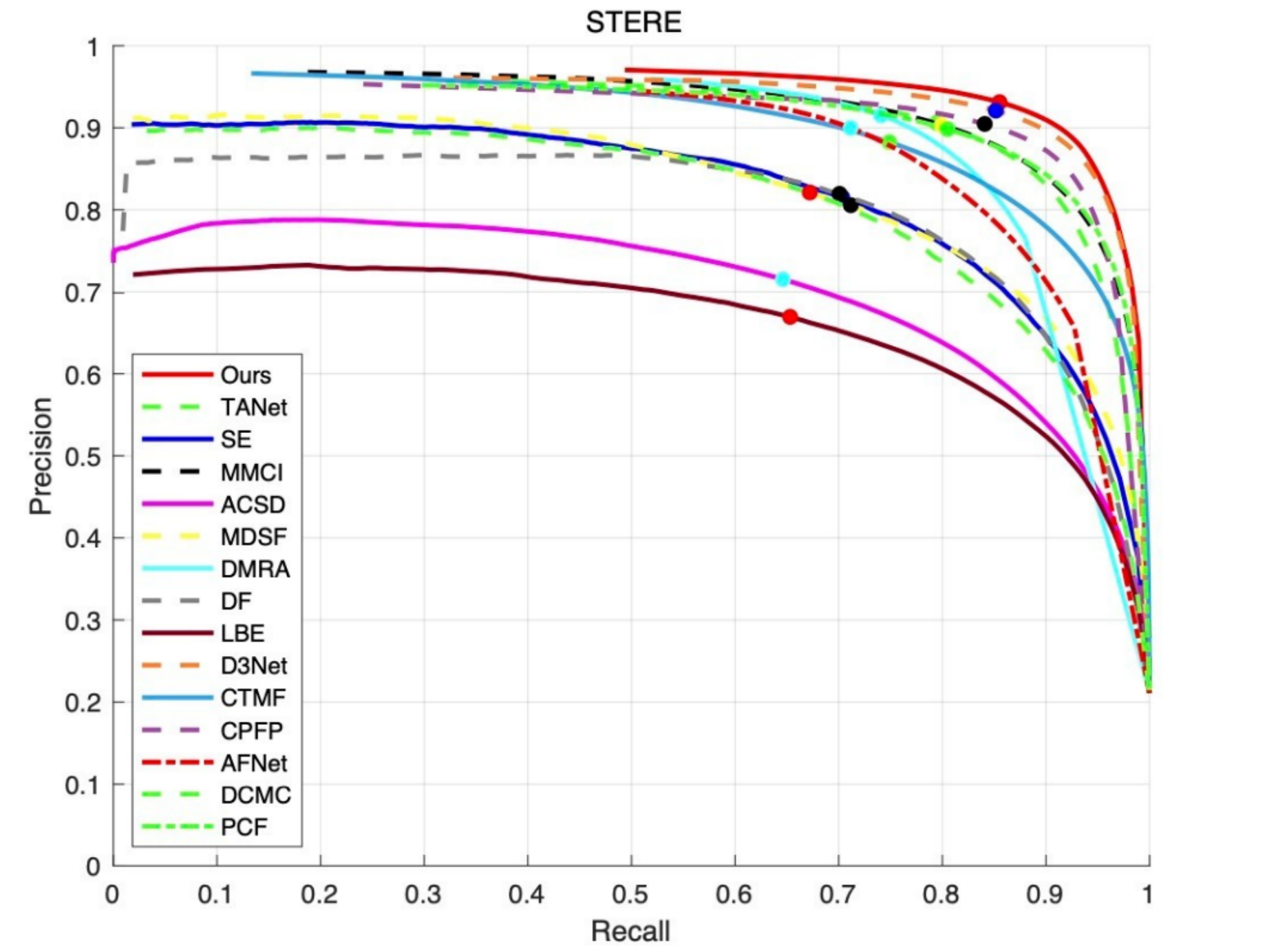}

	}
	\caption{Comparison of PR curves between our method and 14 state-of-the-art methods on 7 datasets. The point on each curve represents the precision and recall used for calculating max F-measure. Compared to other state-of-art methods, our MSIRN method achieves favorable performance on the majority of the data sets.}
     \label{fig7}
\end{figure}

In addition to quantitative experiments, we also present qualitative detection results of different methods on various challenging scenarios. Fig. 8 presents visual comparison between our method and competitive methods including MMCI 
\cite{9chen2019multi}, CTMF 
\cite{57han2017cnns}, DMRA 
\cite{5piao2019depth}, DF 
\cite{64qu2017rgbd}, D3Net 
\cite{50fan2020rethinking}, SE 
\cite{55guo2016salient} and LBE 
\cite{19feng2016local}. Four groups of representative samples are shown in Fig. 8 such as simple scene, low contrast, small object (e.g., the $4^{th}$ and $5^{th}$ rows) and multiple objects. In this figure, our goal is to evaluate whether our MSIRN method could handle various challenging scenarios and produce robust prediction. The qualitative comparisons also confirm the effectiveness of our proposed MSIRN. As the pictures presented in Fig. 8, it shows our method achieves better results on nearly all challenging scenes and generates reasonable saliency maps. In addition, according to the visual comparisons of different scenarios shown in Fig. 8, we can observe that our MSIRN method is more effective to deal with small objects and multiple objects. Take the second row of small object section in Fig. 8 as an example, most of the competitors is confused by the poor depth map. In contrast, our method introduces sufficient cross-modal fusion and eliminates the adverse effects of low quality depth maps. The multiple objects scenario contains co-exists objects with high overlaps. For this challenging situation, most of the methods (e.g. MMCI 
\cite{9chen2019multi} and D3Net 
\cite{50fan2020rethinking}) are not likely to separate each isolated object due to the confusion caused by instance overlap. For example, MMCI 
\cite{9chen2019multi} regards multiple airplanes as one object instead of individual instance. In contrast, our method performs well and segments each object accurately. In addition, our method does not show much advantage than other methods in the simple scene. However, our method is more capable to generate reasonable saliency maps with sharp boundaries in challenging setting such as low contrast.

\begin{figure}[H]
	\centering
	\includegraphics[width=\textwidth]{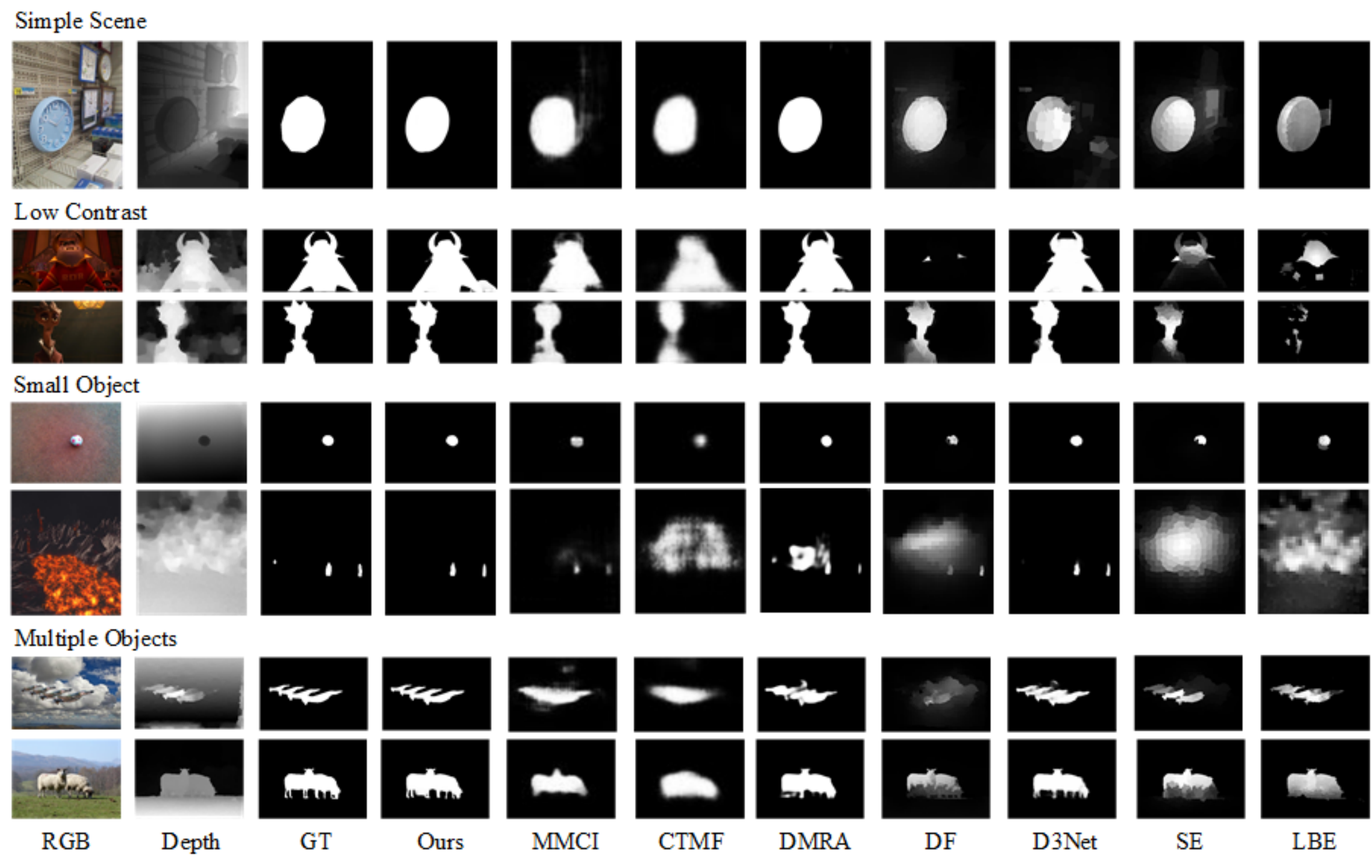}
	\caption{Visual comparison between our method and 7 competitive methods. The input contains challenging scenarios of low contrast (row 2 and 3), small object (row 4 and 5) and multiple objects (row 6 and 7).}
	\label{fig8}
\end{figure}

To sum up, all these results confirm that the model in our proposed method is capable to seperate salient objects from background regions using cross-modal visual cues, and plays more favorable than other competitive methods.

\begin{figure}[H]
	\centering
	\includegraphics[width=\textwidth]{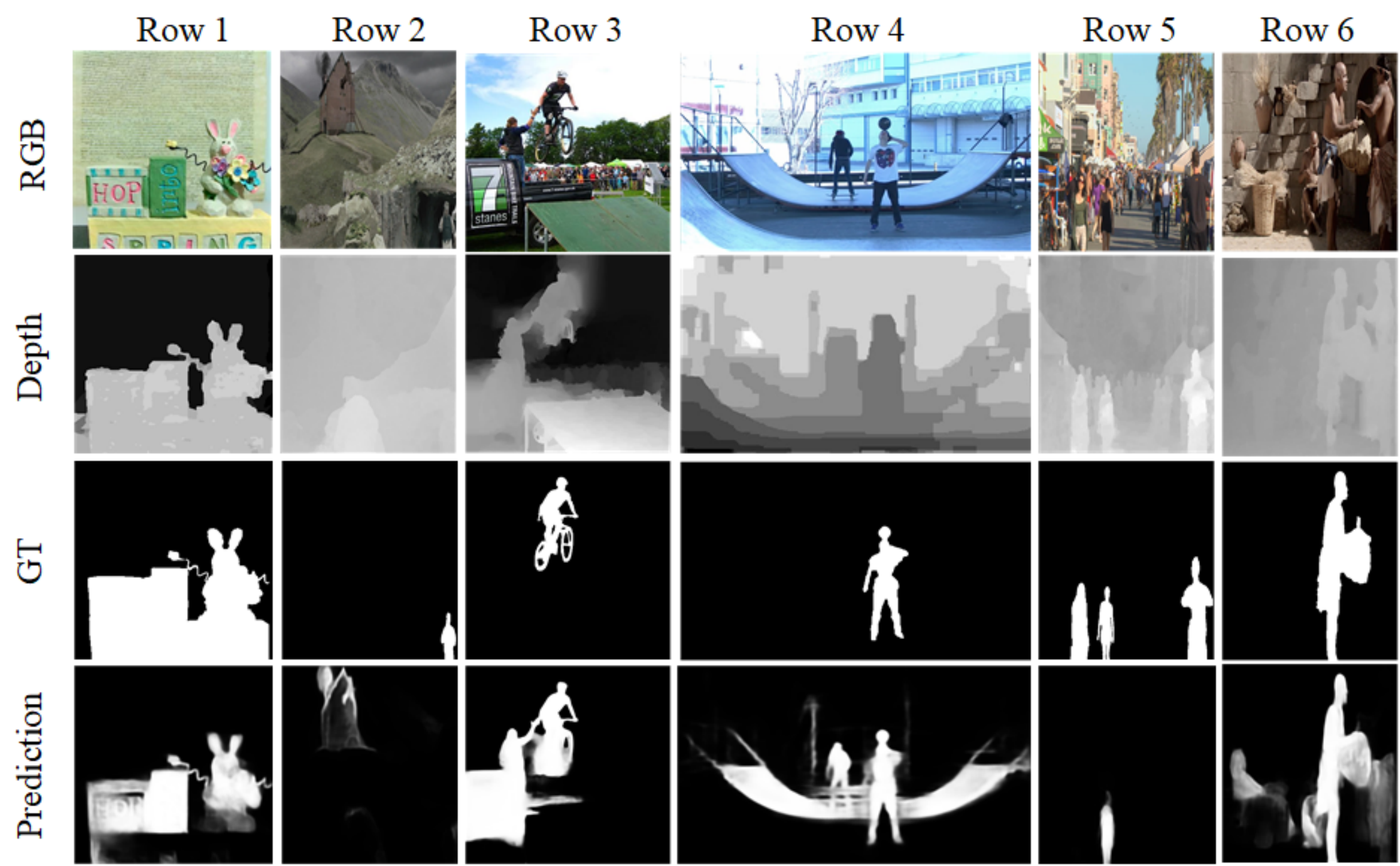}
	\caption{Failure cases of our MSIRN method. In the first 2 columns, the model predicts wrong objects because of the uncertainty of object selection. In the $3^{rd}$  and $4^{th}$ columns, the depth maps provide misleading guidance. In the last 2 columns, the MSIRN is confused by complex scenes.}
	\label{fig8}
\end{figure}

In addition to quantitative and qualitative results, we also report failure cases in Fig. 9. Our method fails to segment salient objects in some extreme conditions. MSIRN exploits multi-modal visual cues in multiple levels. However, complex scenes and coarse depth information affect the detection. Typical examples in Fig. 9 (first and second columns), where our MSIRN locates incorrect salient objects. As shown in the first column of Fig. 9, the model mistakenly regards the table as background object. And in the second column of Fig. 9, the model incorrectly selects the house as salient object instead of the small person in the corner of the image. The reason for the second situation (e.g. the $3^{rd}$  and $4^{th}$ columns in Fig. 9) is that the depth maps provide misleading information. The depth signal including spatial layout of objects provides incorrect guidance for salient detection. The last situation (e.g. the last two columns in Fig. 9) which generates incorrect results when the model encounters complex scenes such as crowded street. The root cause for this failure is that, co-occurrence of objects confuses the detector. As illustrated in the 5th column in Fig. 9, the model regards the single person in the middle as salient object instead of multiple people.

\begin{table}[H]
  \centering
  \caption{Ablation studies for the proposed method on seven datasets. The coarse refinement (CR), channel-wise attention (CA), spatial attention (SA), atrous spatial pyramid pooling (ASPP), fine-grained refinement (FR), feature aggregation (FA) and skip connection (SC) are removed from MSIRN.}
    \begin{tabular}{cc|cccc|ccc|c}
    \hline
    \hline
    \multicolumn{2}{c}{} & \multicolumn{4}{|c|}{Coarse branch}&\multicolumn{3}{c|}{Fine branch} & \multicolumn{1}{c}{MSIRN} \\
    \cline{3-10}
 &Metric&w/o CR&w/o CA&w/o SA&w/o ASPP&w/o FR&w/o FA&w/o SC&Ours \\
 & &(a)&(b)&(c)&(d)&(e)&(f)&(g)& \\
    \hline
    \hline
\multirow{3}*{\rotatebox{90}{\emph{NJU2K}}} &$S_\alpha$↑ &0.902 & 0.909 & 0.912 & 0.910 & 0.900 & \textbf{0.913} & 0.909 & \textbf{0.912} \\
 &$F_\beta$↑   &0.886 & 0.889 & 0.894 & 0.884 & 0.876 & 0.884          & 0.901 & \textbf{0.904} \\
&$\mathcal{M}$↓   &0.042 & 0.043 & 0.041 & 0.043 & 0.047 & 0.044          & 0.039 & \textbf{0.038}\\

\hline
\multirow{3}*{\rotatebox{90}{\emph{LFSD}}} &$S_\alpha$↑ &0.850 & 0.854 & 0.855 & 0.860 & 0.850 & 0.851 & \textbf{0.862} & \textbf{0.862} \\
 &$F_\beta$↑   &0.816 & 0.826 & 0.835 & 0.827 & 0.816 & 0.822 & \textbf{0.842} & 0.838          \\
&$\mathcal{M}$↓   &0.085 & 0.082 & 0.079 & 0.081 & 0.091 & 0.088 & \textbf{0.074} & 0.077         \\

\hline
\multirow{3}*{\rotatebox{90}{\emph{STERE}}} &$S_\alpha$↑ &0.903 & 0.906 & 0.903 & 0.897 & 0.897 & 0.903 & \textbf{0.904} & \textbf{0.904} \\
 &$F_\beta$↑   &0.865 & 0.872 & 0.875 & 0.851 & 0.849 & 0.863 & 0.877          & \textbf{0.892} \\
&$\mathcal{M}$↓   &0.049 & 0.045 & 0.044 & 0.053 & 0.054 & 0.050 & \textbf{0.043} & 0.045         \\

\hline
\multirow{3}*{\rotatebox{90}{\emph{DES}}} &$S_\alpha$↑ &0.929 & 0.930          & 0.920 & 0.929 & 0.917 & 0.925 & 0.926 & \textbf{0.931} \\
 &$F_\beta$↑ &0.896 & 0.896          & 0.896 & 0.882 & 0.873 & 0.887 & 0.899 & \textbf{0.903} \\
&$\mathcal{M}$↓   &0.025 & \textbf{0.023} & 0.025 & 0.026 & 0.029 & 0.027 & 0.024 & \textbf{0.023}\\

\hline
\multirow{3}*{\rotatebox{90}{\emph{NLPR}}} &$S_\alpha$↑ &0.929 & 0.923 & 0.930          & 0.923 & 0.917 & \textbf{0.931} & 0.928          & \textbf{0.931} \\
 &$F_\beta$↑ &0.883 & 0.880 & 0.891          & 0.874 & 0.862 & 0.884          & 0.893          & \textbf{0.894} \\
&$\mathcal{M}$↓   &0.027 & 0.028 & \textbf{0.025} & 0.029 & 0.032 & 0.027          & \textbf{0.025} & \textbf{0.025}\\

\hline
\multirow{3}*{\rotatebox{90}{\emph{SSD}}} &$S_\alpha$↑ &0.873 & 0.879          & 0.871 & 0.871 & 0.876 & 0.879 & 0.878 & \textbf{0.880} \\
 &$F_\beta$↑ &0.823 & 0.839          & 0.829 & 0.818 & 0.822 & 0.840 & 0.836 & \textbf{0.843} \\
&$\mathcal{M}$↓   &0.055 & \textbf{0.049} & 0.050 & 0.054 & 0.056 & 0.051 & 0.051 & 0.050          \\

\hline
\multirow{3}*{\rotatebox{90}{\emph{SIP}}} &$S_\alpha$↑ &0.866 & 0.861          & 0.868 & 0.852 & 0.862 & 0.864 & 0.865 & \textbf{0.879} \\
 &$F_\beta$↑ &0.846 & 0.849          & 0.857 & 0.823 & 0.840 & 0.848 & 0.859 & \textbf{0.867} \\
&$\mathcal{M}$↓   &0.066 & 0.066          & 0.061 & 0.074 & 0.069 & 0.066 & 0.062 & \textbf{0.056}\\

    \hline
    \end{tabular}
  \label{tab:addlabel}
\end{table}%

\subsection{Ablation Study}

In this subsection, our goal is to investigate the contribution of each component of our proposed method. The evaluated contribution includes i) the benefits of coarse refinement as well as fine-grained refinement for our proposed MSIRN; ii) the effectiveness of ASPP layer, channel-wise and spatial attention for cross-modal fusion component; iii) the improvements of feature aggregation layer and skip connection for fine-grained refinement.

\textbf{Effects of coarse refinement (CR) and fine-grained refinement (FR).} In order to verify the effectiveness of our proposed coarse refinement as well as fine-grained refinement for our proposed method, we conduct ablation studies by removing specific module (e.g., remove coarse refinement supervision or fine-grained refinement from original network). The networks without coarse refinement and fine-grained refinement are “w/o CR” and “w/o FR” respectively. In Table 4, by removing coarse refinement in our network, the performance decreases to some extent ($S_\alpha$ 0.912 vs. 0.902 on \emph{NJU2K}). On the other hand, when removing fine-grained refinement and using the 6th refinement saliency map as output, the performance drops sharply ($S_\alpha$ 0.912 vs. 0.900 on \emph{NJU2K}), which provides evidence for the effectiveness of refinement. These effects are also reflected in Fig. 10. The foreground salient objects are detected accurately with the aid of coarse-to-fine refinement; however, without iterative refinement, the model tends to produce inferior results due to the lack of potential guidance. For example, as for the red house in the third column, the model without iterative refinement cannot find complete foreground objects as lack of exploration of potential features. When we add two types of refinements into our network, the performance boosts 1.8\% and 2.8\% on \emph{NJU2K} in terms of $F_\beta$ for coarse and fine-grained refinement respectively. The visual results in Fig. 11 also give evidence to the benefit of our proposed iterative refinement.

\begin{figure}[H]
	\centering
	\includegraphics[width=0.4\textwidth]{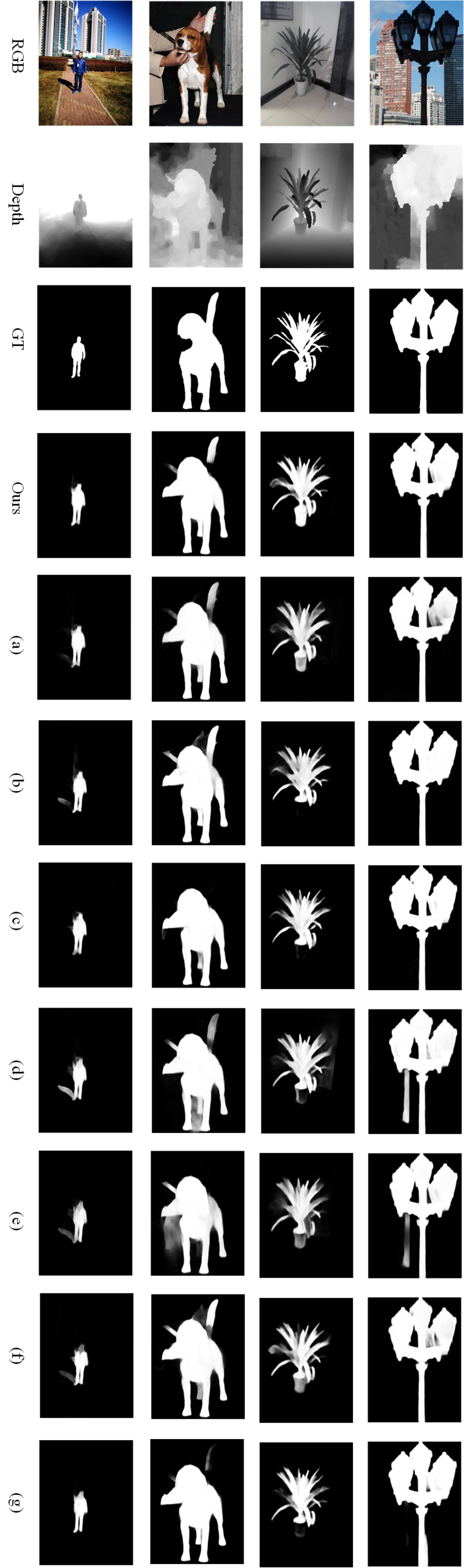}
	\caption{The visual comparison of ablation studies. The index of ablation variant is shown in Table 4. Row (a) – (g) denote predicted saliency map without coarse refinement (CR), channel-wise attention (CA), spatial attention (SA), atrous spatial pyramid pooling (ASPP), fine-grained refinement (FR), feature aggregation (FA) and skip connection (SC) module, respectively.}
	\label{fig10}
\end{figure}

As can be seen in Fig. 11, iterative refinement provides steady improvements to each level saliency maps produced by our MSIRN network.

\begin{figure}[H]
	\centering
	\includegraphics[width=\textwidth]{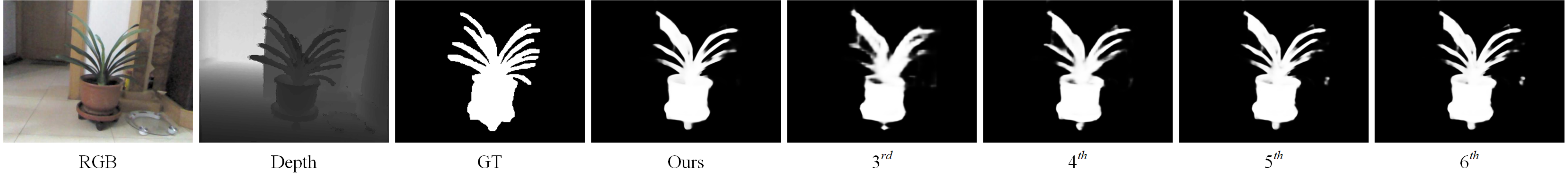}
	\caption{The predicted saliency maps from $3^{rd}$ to $6^{th}$ refinement (Column 5 to 8) as well as the last refinement (Column 4) resulting from fine-grained refinement branch.}
	\label{fig11}
\end{figure}

\textbf{Effect of ASPP layer, channel-wise (CA) and spatial attention (SA).} We conduct ablation studies to verify the contributions of channel-wise attention, spatial attention as well as ASPP layer. We remove the ASPP layer, channel-wise, spatial attention and construct three variants,which are denoted as “w/o ASPP”, “w/o CA” and “w/o SA” respectively. Comparing the original results of MSIRN with “w/o ASPP”, “w/o CA” and “w/o SA” in Table 4, we can see  2\%,  1.5\%,  1\% performance decrease in terms of $F_\beta$ on \emph{NJU2K} dataset. As shown in Fig. 10, adding both attention and ASPP layers can better locate salient object and produce sharper result in fine detail. Meanwhile, as shown in Fig. 11, the predictions from MSIRN contain more complete information and sharp saliency maps compared with three ablation variants. This observation confirms the effectiveness of our attention based cross-modal fusion method. The proposed attention layer is beneficial to highlight complementary cross-modal information and suppress unnecessary features.

\textbf{Effect of feature aggregation (FA) and skip connection (SC).} To illustrate the effectiveness of our method, we construct two variants of our model by removing feature aggregation and skip connection layers, e.g. “w/o FA” and “w/o SC”. As shown in Table 4, there is a decrease when removing either feature aggregation (FA) or skip connection (SC) layer, e.g., 0.904 vs. 0.884, 0.904 vs. 0.901 in terms of $F_\beta$ on \emph{NJU2K} dataset. As for visual comparison in Fig. (f) and Fig. (g), the MSIRN result contains more details and achieves better performance. This indicates that it is necessary to ultilize larger receptive field and skip connection to enhance feature representation.

\textbf{Effect of K in Coarse Refinement.} In this subsection, our goal is to evaluate the effects of supervision steps in coarse refinement branch. We gradually increase the number of supervision steps. As shown in Fig. 12, increasing the number of supervision steps leads to steady improvement, but the computation cost and responsing time are also increased. Besides, reducing refinement steps degrades the feature extracting process and generates inferior results. However, the selection of refinement steps should balance between accuracy and computational cost in real-world application. To achieve optimal balance, we select refinement step to 6 in coarse refinement branch. In terms of speed, our network achieves 12 fps and 42 fps accelerated by a single 1080ti GPU with minibatch of 1 and 8 respectively.

\begin{figure}[H]
	\centering
	\includegraphics[width=\textwidth]{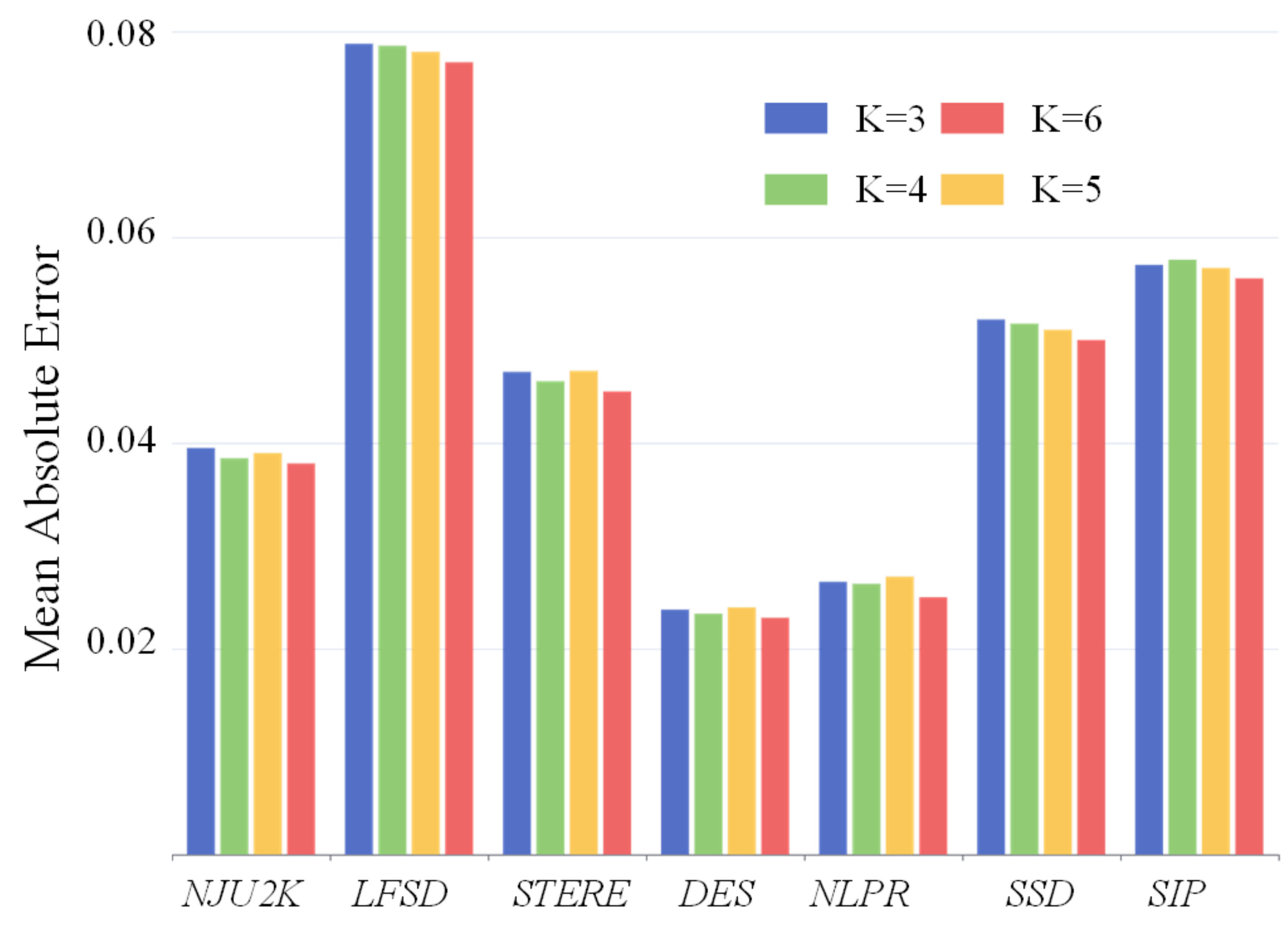}
	\caption{Influence of different K parameters in coarse refinement branch on seven datasets.}
	\label{fig12}
\end{figure}

\section{Conclusions}

In this paper, we develop a novel paradigm for extracting salient objects on RGB-D multi-modal input. The goal of our method is to learn a fine-grained model, which can segment the foreground salient object in fine detail reinforced by both RGB and depth streams. In order to achieve this goal, this study attempts to define an update model in which an iterative refinement pipeline is proposed. Complementary multi-level information is leveraged in a coarse-to-fine iterative refinement process. Shallow layers of bottom convolution give guidance to object boundaries and local details, while deep layers of top convolution provide semantic information which helps to refine saliency targets. To fully exploit and dynamically fuse multi-modal visual cues in detection, we develop a strategy with both channel-wise and spatial multiplication to obtain a feature representation simultaneously escalating cross-modal learning and eliminating misleading information. In this design, channel-wise and spatial attention as well as ASPP layer are leveraged to learn the visual representation of multi-scale features in cross-model learning, which is the key learning step to achieve good performance for salient prediction. After the coarse salient prediction branch, we further employ an encoder-decoder based fine-grained refinement branch. Feature aggregation modules with skip connections are introduced in encoder stage to enlarge receptive fields. Experimental evaluations on seven challenging datasets validate the advantage of our proposed method comparied with 15 SOTA methods.

In the future, we plan to integrate transformer module to extract the co-occurring visual cues in both RGB and depth streams. For application, MSIRN will be applied to background changing scenario and camouflaged object detection.

\nocite{*}

\bibliography{bibfile}
\end{document}